\documentclass[lettersize,journal]{IEEEtran}
\usepackage{times}

\usepackage[numbers]{natbib}
\usepackage{multicol}
\usepackage{acronym}
\usepackage{capt-of}
\usepackage{mwe}
\usepackage[bookmarks=true]{hyperref}
\usepackage{float}   
\usepackage{booktabs}  
\usepackage{xcolor}
\usepackage{makecell}
\usepackage{amsmath}
\usepackage[clock]{ifsym}
\usepackage{multirow}

\newcommand{\subsubsubsection}[1]{%
  \par\textbf{#1.}\quad%
}

\begin{document}

\acrodef{EMG}{electromyography}
\acrodef{HDEMG}{high-density electromyography}
\acrodef{cSCI}{cervical spinal cord injury}
\acrodef{ADLs}{activities of daily living}
\acrodef{DOFs}{degrees of freedom}
\acrodef{ML}{machine learning}
\acrodef{CNN}{convolutional neural network}
\acrodef{RMS}{root-mean-square}
\acrodef{MVC}{maximum voluntary contraction}
\acrodef{SUS}{system usability score}

\title{Bimanual High-Density EMG Control for In-Home Mobile Manipulation by Users with Quadriplegia}

\author{
\authorblockN{
Jehan Yang,
Eleanor Hodgson,
Cindy Sun,
Zackory Erickson\textsuperscript{\textdagger},
and Douglas J. Weber\textsuperscript{\textdagger}
}\\
\authorblockA{
Carnegie Mellon University\\
Email: jehany@andrew.cmu.edu
}
\authorblockA{
\textsuperscript{\textdagger}Equal advising
}
}


%

\makeatletter
\let\@oldmaketitle\@maketitle
\renewcommand{\@maketitle}{\@oldmaketitle
    \centering
    \begin{minipage}[t]{0.34\linewidth}
        \centering
        \includegraphics[width=\linewidth]{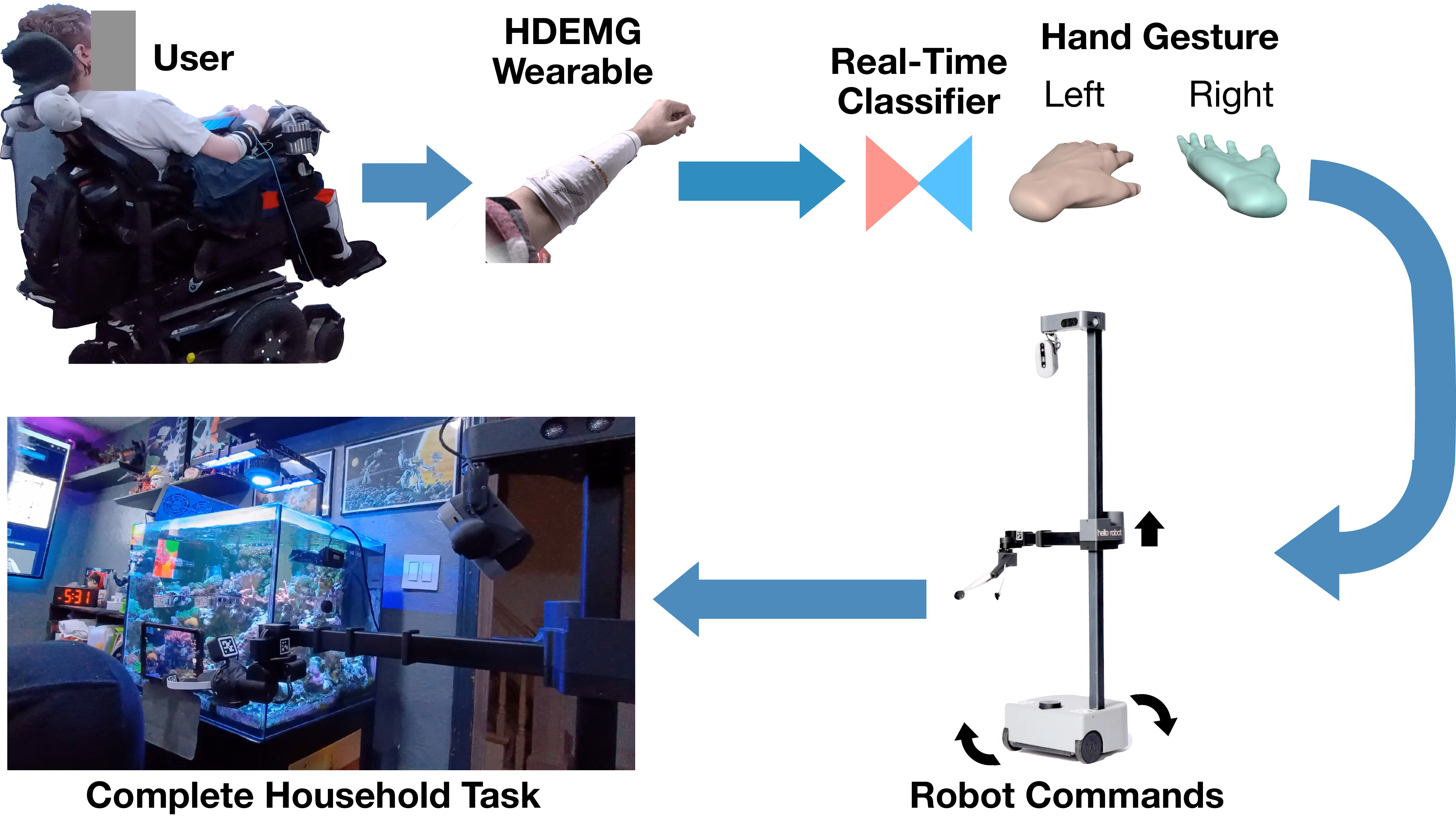}
    \end{minipage}\hfill
    \begin{minipage}[t]{0.64\linewidth}
        \centering
        \includegraphics[width=\linewidth]{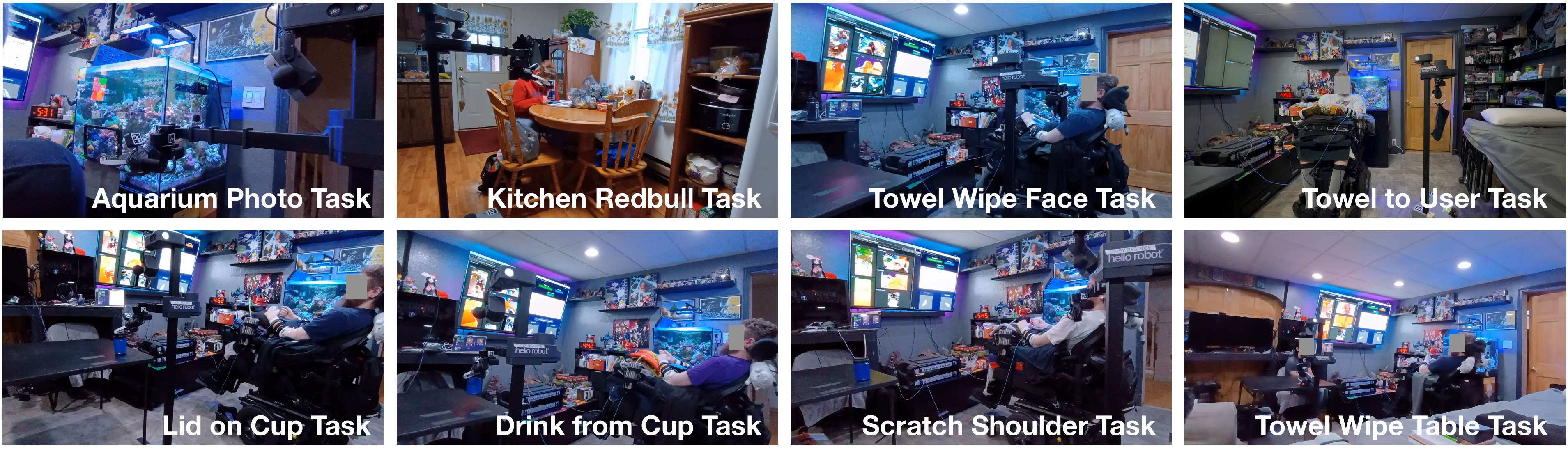}
    \end{minipage}
    \vspace{-2mm}
  \small
    \captionof{figure}{(Left) System overview showing a user with cSCI and limited wrist movements. \ac{EMG} signals are classified on each forearm to infer user intent without requiring overt hand motion, and mapped to robot actions. (Right) Tasks performed in the user’s home during the finalized study phase.}
    \label{fig:system-overview}
  \vspace{-8.5mm}
  }
\makeatother

\maketitle

\begin{abstract}
Mobile manipulators in the home can enable people with \ac{cSCI} to perform daily physical household tasks that they could not otherwise do themselves. However, paralysis in these users often limits access to traditional robot control interfaces such as joysticks or keyboards. In this work, we introduce and deploy the first system that enables a user with quadriplegia to control a mobile manipulator using intent from paralyzed parts of their body, using bimanual \ac{HDEMG}. We develop a pair of custom, fabric-integrated \ac{HDEMG} forearm sleeves, worn on both arms, that capture residual neuromotor activity from clinically paralyzed degrees of freedom and support real-time gesture-based robot control. We achieve high classification accuracies based on motor intent across (\(\boldsymbol{n=2}\)) users with \ac{cSCI}, achieving up to \(\mathbf{98.0\%}\). Second, by integrating vision, language, and motion planning modules, we introduce a shared autonomy framework that supports robust and user-driven teleoperation, with particular benefits for navigation-intensive tasks in home environments. Finally, to demonstrate the system in the wild, we present a twelve-day in-home user study evaluating real-time use of the wearable EMG interface for daily robot control. Together, these system components enable effective robot control for performing \ac{ADLs} and other household tasks in a real home environment.
\vspace{-1em}
\end{abstract}

\IEEEpeerreviewmaketitle
\addtocounter{figure}{-1}

\section{Introduction}
\label{sec:introduction}
Millions of people around the world suffer from quadriplegia, a condition in which much of the movement and sensation of the trunk, upper limbs, and lower limbs are lost. One of the most common causes of quadriplegia is \ac{cSCI}~\cite{armour2016prevalence}. Due to the loss of trunk and arm movements, many people with \ac{cSCI} lose the ability to perform many \ac{ADLs}~\cite{hammell2004exploring}, particularly those requiring reaching and grasping. These tasks include feeding, drinking, mobility, and hygiene~\cite{li2017longitudinal}, making users highly dependent on caregivers.

Mobile manipulators, such as the Hello Robot Stretch~\cite{kemp2022design}, enable physical assistive manipulation throughout the home, whether the user is sitting in a wheelchair or lying in bed~\cite{padmanabha2024independence}. A key challenge to enable mobile manipulation for users with quadriplegia involves designing expressive and accessible control interfaces. Although users with \ac{cSCI} may have little to no visible movement in their hands, previous work has demonstrated that \ac{EMG} signals can capture intended hand and finger gestures even from clinically paralyzed degrees of freedom~\cite{despradel2026enabling, ting2021sensing, yang2025intuitive, yang2025non, oliveira2024direct}. These signals arise from residual motor neuron activity and motor unit action potentials that persist despite the absence of overt motion. When combined with machine learning methods and \ac{HDEMG}, these residual patterns can be reliably detected using sensors placed on the forearm, enabling real-time classification of intended gestures with minimal skin preparation and without precise electrode placement. This capability opens the door to wearable, hands-free control interfaces that leverage biomechanical degrees of freedom that are not available in traditional physical controllers.

Although prior work has demonstrated decoding residual neuromotor signals from the forearm in individuals with spinal cord injury, these efforts have mainly focused on offline decoding, laboratory settings, or single-arm sensing configurations~\cite{despradel2026enabling, ting2021sensing, yang2025intuitive, yang2025non, oliveira2024direct}. In contrast, we introduce the first fabric-integrated bimanual \ac{HDEMG} forearm sleeve designed specifically for real-time robot control, enabling simultaneous intent decoding from both arms during mobile manipulation. By distributing 128 electrodes per forearm within a wearable textile form factor, our design has the advantage of residual neuromotor signal coverage, as well as comfort and rapid donning.

However, directly teleoperating a mobile manipulator using low-dimensional interfaces remains a challenge. Most teleoperation systems require mapping one- or two-degree-of-freedom inputs to high-dimensional robot motion, increasing cognitive load and susceptibility to errors. 
In this work, we combine wearable neuromotor control and shared autonomy to enable practical, user-driven mobile manipulation in the home. An overview of the complete sensing, classification, and control pipeline is shown in Figure~\ref{fig:system-overview}.
Our system integrates an \ac{HDEMG}-based gesture interface with a shared autonomy framework that incorporates open-vocabulary vision--language grounding for object alignment, room-level navigation assistance, and LiDAR-based collision safety. This design allows the user to retain high-level control over robot behavior while the system assists with perception, alignment, and safety in an unstructured home environment.

We evaluate the complete system through a 12-day longitudinal deployment in the home with a user with complete motor and sensory quadriplegia from \ac{cSCI}. The study consists of a five-day exploratory phase used to collect \ac{EMG} data, train and refine real-time gesture classifiers, and adapt the teleoperation interface, followed by a seven-day standardized study phase using a fixed system configuration. Throughout the deployment, the user performed a diverse set of tasks, including drinking, opening doors, wiping surfaces, retrieving objects, and positioning a tablet to take photos of an aquarium. These tasks span basic ADLs, instrumental ADLs, and personalized activities, reflecting both independence-critical and quality-of-life-oriented use cases.

In summary, this paper makes three primary contributions:
\begin{itemize}
    \item We introduce the \textbf{first bimanual fabric-integrated high-density EMG (HDEMG) forearm sleeves designed for real-time robot control}, enabling a user with quadriplegia to control a robot using intended movements from paralyzed parts of their body, while open-sourcing the hardware design. 
    \item We present a \textbf{shared autonomy teleoperation framework} that integrates perception, navigation, and safety to support robust mobile manipulation in the home while preserving user control authority.
    \item We demonstrate the \textbf{first in-home deployment of wearable HDEMG for real-time mobile manipulator control by a quadriplegic user}, evaluated in a 12-day longitudinal study in 11 tasks.
\end{itemize}


\section{Related Work}

\subsection{Control Interfaces for Assistive Teleoperation}
A wide range of control interfaces for robot teleoperation have been explored, including joysticks, button-based controllers, and customizable web-based interfaces~\cite{bjornfot2021evaluating,ranganeni2023evaluating,wojtowicz2023stretch}. Higher-bandwidth spatial input devices, such as VR controllers and scaled robot arm twins, have enabled expressive teleoperation for able-bodied users~\cite{hu2025rac,wu2024gello,fu2024mobile}, but remain largely inaccessible to individuals with quadriplegia.

As a result, users with quadriplegia commonly rely on alternative interfaces such as sip-and-puff systems and mouth-operated joysticks~\cite{da2019development,huo2008introduction}. Brain--computer interfaces have also been explored, although many approaches require invasive or semi-invasive sensing to support continuous manipulation~\cite{hughes2020bidirectional}. These limitations motivate non-invasive wearable sensing modalities that can capture user intent without requiring overt motion or fine motor control. Compared to head-worn interfaces such as HAT~\citep{padmanabha2022hat}, EMG-based control does not require dedicating head motion to robot control, which may better preserve gaze flexibility and situational awareness during teleoperation, and can support users with limited or fatiguing head mobility.

\subsection{EMG-Based Neuromotor Interfaces}
EMG-based interfaces provide a non-invasive means of sensing neuromotor intent by measuring muscle activity associated with motor neuron activation, including residual signals present in individuals with paralysis~\cite{oliveira2024direct}. Previous work has demonstrated reliable decoding of gesture-related EMG signals in individuals with quadriplegia, including signals corresponding to clinically paralyzed degrees of freedom~\cite{ting2021sensing,despradel2026enabling,yang2025non}. Recent studies have highlighted the importance of large-scale datasets and adaptation strategies to improve the robustness to variability between sessions and electrode shift~\cite{kaifosh2025generic,du2017surface,yang2024emgbench}. Although effective, many robustness and adaptation strategies require large datasets of the target population or extended training, motivating interface designs that prioritize rapid daily calibration and stability under practical constraints in home.

Wearable EMG interfaces have previously been implemented in rigid wristbands~\cite{sathiyanarayanan2016myo,kaifosh2025generic} and in HDEMG sleeves based on a single-arm fabric for gesture recognition in post-stroke populations~\cite{meyers2024decoding}, although without reporting on fabrication or long-term deployment. Wearable HDEMG has also been used to control mobile manipulators in laboratory settings with able-bodied users~\cite{yang2025high}. In contrast, the present work introduces the first bimanual, fabric-integrated HDEMG sleeve system for robot control and evaluates the first EMG-based robot interface deployed in a real home environment by a user with motor and sensory-complete quadriplegia. 

\subsection{Shared Autonomy and In-Home Robot Deployment}
Shared autonomy has been widely studied as a means of reducing the teleoperation burden by combining user input with autonomous robot capabilities~\cite{padmanabha2024independence,tao2024incremental,losey2022learning}. Approaches range from goal-oriented assistance and input blending to safety-oriented action filtering and automatic mode switching~\cite{nakamura2025generalizing,tao2025lams}. Although these methods have shown promise, much of the literature has been evaluated in laboratory or semi-structured environments, where failures are inexpensive, and system assumptions can be tightly controlled.

Despite significant progress in robot manipulation, relatively few systems have been deployed for repeat in-home use. Wheelchair-mounted robotic arms such as the Kinova Jaco have been successfully adopted by many users and provide meaningful daily assistance~\cite{styler2025qualitative}, but their fixed mounting and workspace can limit flexibility in locations and postures~\cite{hutmacher2025identification}. Some recent studies have reported multi-day home deployments with users with motor impairments using manipulators and mobile manipulators~\cite{padmanabha2024independence,nanavati2025lessons,jenamani2025feast}, but none provide longitudinal evaluations of various daily tasks controlled by wearable neuromotor interfaces allowing control of a robot using paralyzed degrees of freedom of the user.

Our work contributes to this space by presenting a longitudinal in-home deployment of a mobile manipulator controlled via a wearable HDEMG interface, evaluated across a broad set of daily living, instrumental, and personalized tasks.

\begin{figure}[t!]
    \centering
    \includegraphics[width=0.9\linewidth]{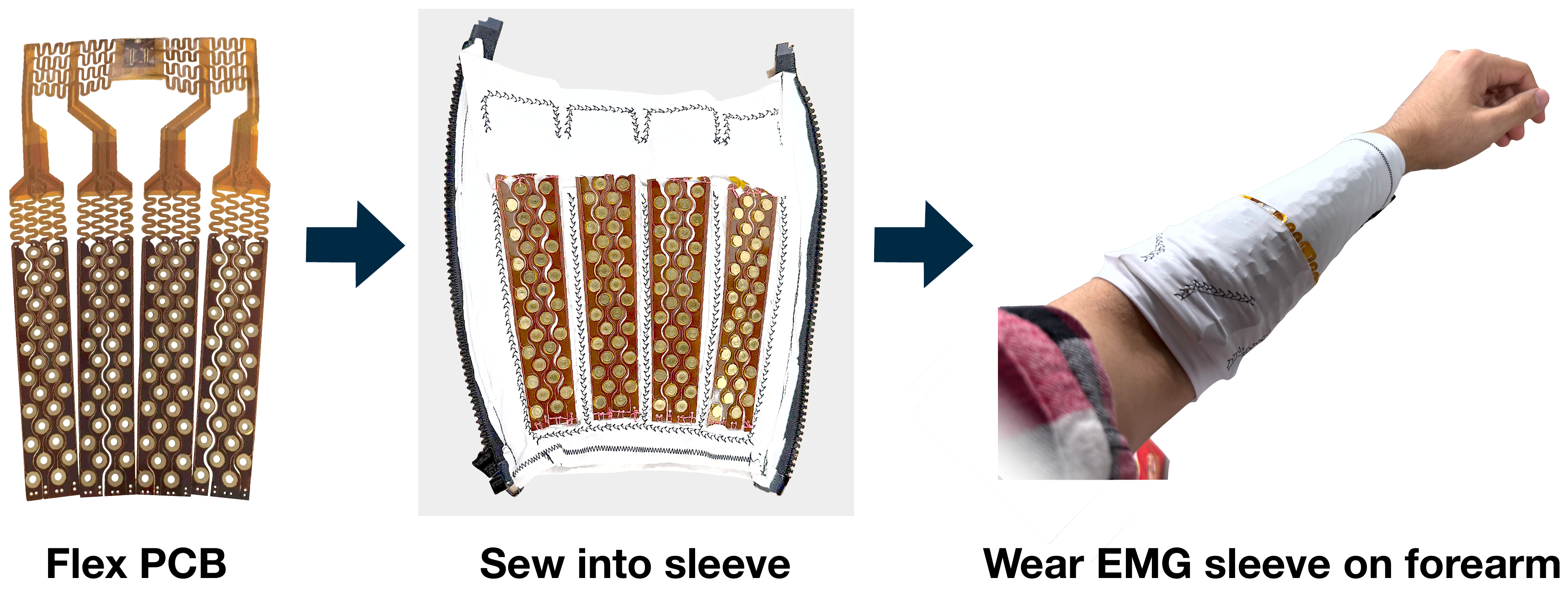}
    \caption{Flex PCB has 128 gold-plated electrodes soldered onto pads, folded and sewn into spandex fabric forearm sleeve, and attached onto both arms of the user. Three Velcro strips are also attached per sleeve in order to ensure electrodes make tight contact with the arm.}
    \label{fig:flex-pcb}
\end{figure}

\begin{figure*}
    \centering
    \includegraphics[width=\linewidth]{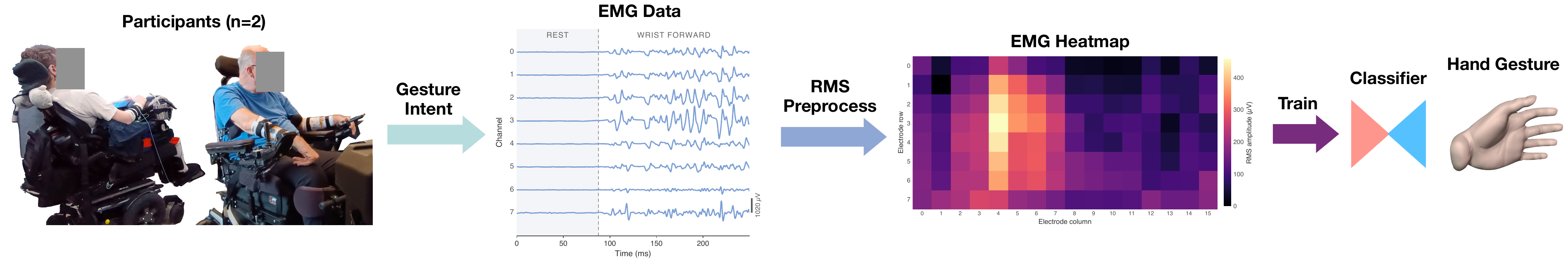}
    \vspace{-6.5mm}
  \small
    \caption{We collected EMG data from two users with quadriplegia to evaluate whether our hardware system and machine learning pipeline can accurately classify hand gestures. We selected an RMS heatmap for preprocessing and CNN architecture for classification following a sweep across different architectures using data from a screening study.}
    \label{fig:data-collection-two-users}
\end{figure*}

\section{Wearable HDEMG Interface}
\subsection{Wearable Bimanual HDEMG Sleeves}
\label{sec:wearable-hdemg}
We introduce the first bimanual high-density EMG full-forearm sleeve integrated into a fabric wearable for robot control, designed to support extended daily use and simultaneous intent decoding from both arms. The sleeve is constructed using a textile substrate sewn using a standard sewing machine into which a flexible printed circuit board (PCB) is integrated. Each sleeve contains 128 surface electrodes per arm, arranged in an 8$\times$16 grid to provide dense spatial coverage throughout the forearm. This configuration enables high-resolution sensing of residual neuromotor activity while maintaining a low-profile, wearable form factor suitable for home deployment.

The flexible PCB is designed with rigid electrode ``fingers'' that extend longitudinally along the forearm, while serpentine traces are used between fingers to provide mechanical compliance. Inner and outer views of the sleeve layout, including electrode placement and serpentine interconnects, are shown in Supplementary Figures \ref{supplementary:sleeve-inner-view} and \ref{supplementary:sleeve-outer-view}, respectively. This hybrid rigid--flexible structure allows the sleeve to conform to the user’s arm while preserving consistent electrode spacing and contact. The serpentine interconnects improve the coverage of muscles distributed circumferentially around the forearm and reduce stress during donning, doffing, and arm movement. For the users in this study, the sleeve provided coverage throughout approximately a 27~cm forearm circumference and a 24~cm forearm length, allowing detection of both superficial and deeper muscle groups relevant for gesture intent decoding.

Electrodes are distributed along fingers of 12~cm in length, with an effective circumferential coverage of at least 16~cm, with the ability to stretch through the use of Spandex fabric and serpentine traces. This geometry was chosen to accommodate a wide range of arm sizes and shapes, particularly for users with muscle atrophy resulting from spinal cord injury. Using spatially-dense HDEMG instead of precise anatomical placement as is typical in clinical studies with \ac{EMG}, the sleeve reduces sensitivity to exact positioning and enables robust gesture classification without extensive skin preparation or electrode alignment, supporting rapid daily setup in a home environment. A standard alcohol wipe is used to clean the electrodes between sessions.

\begin{figure*}
    \centering
    \begin{minipage}[t]{0.23\linewidth}
        \centering
        \includegraphics[width=\linewidth]{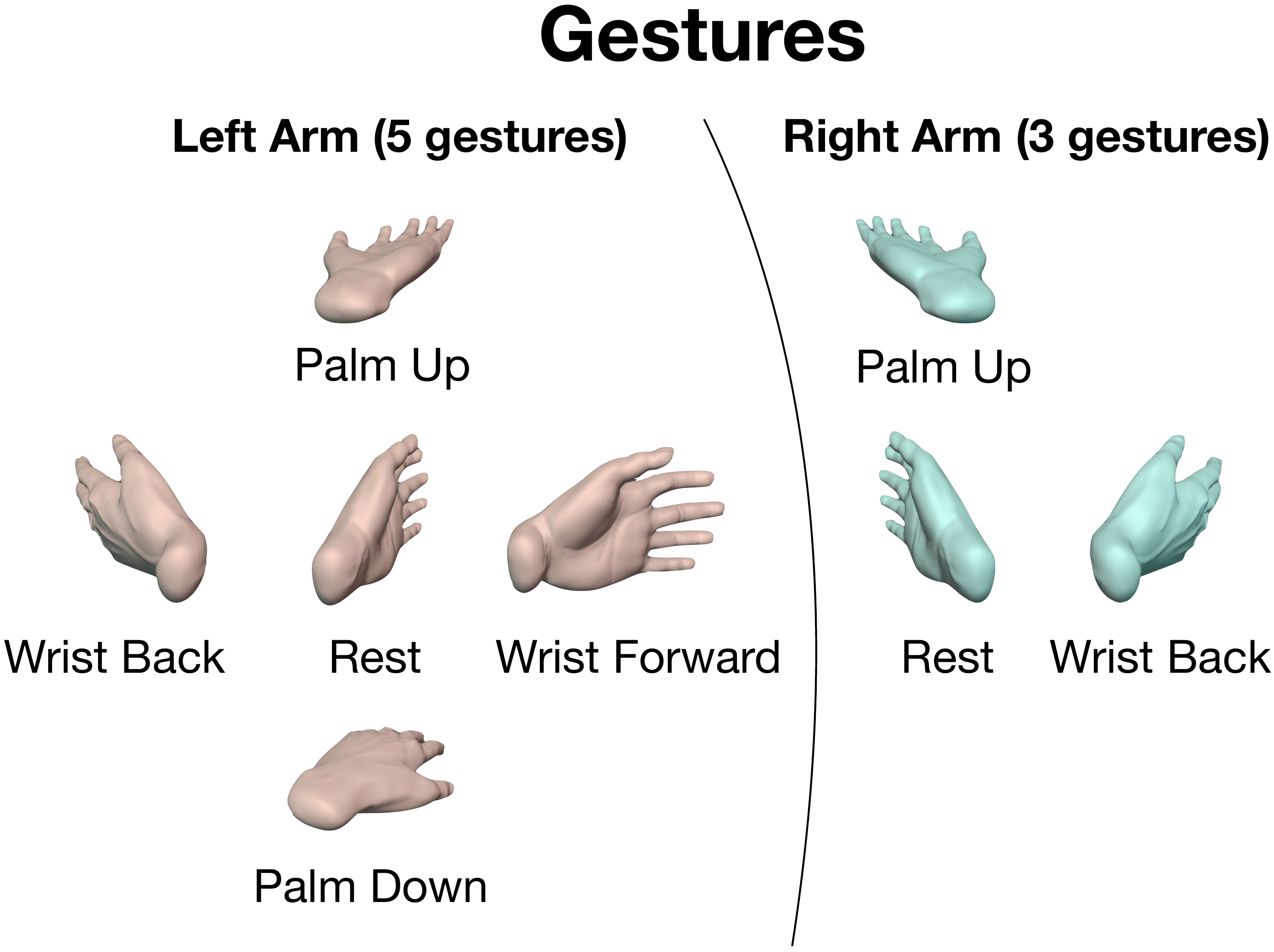}
    \end{minipage}\hfill
    \begin{minipage}[t]{0.73\linewidth}
        \centering
        \includegraphics[width=\linewidth]{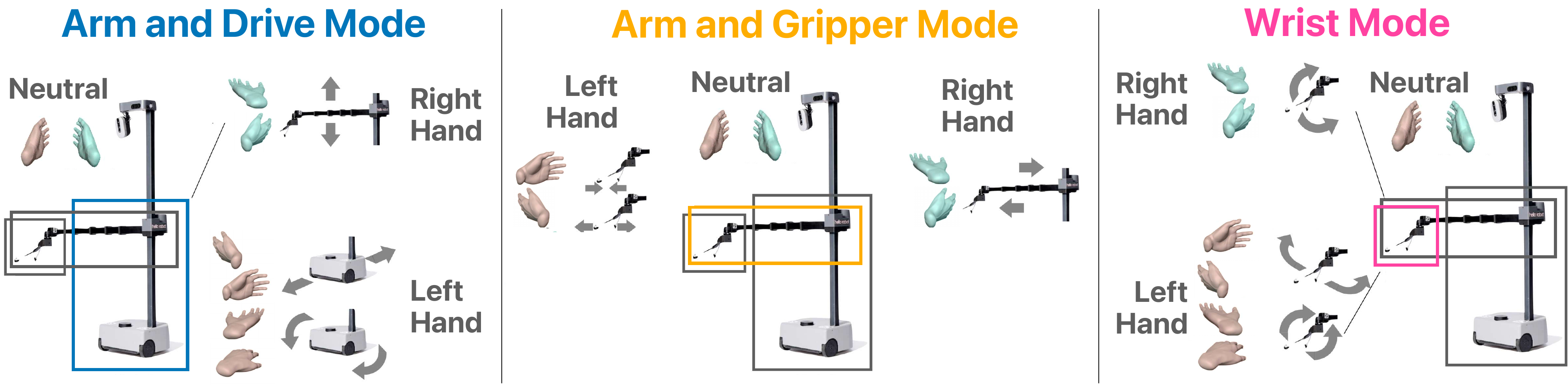}
    \end{minipage}
  \small
    \captionof{figure}{(Left) Gestures used for robot control on both hands for the last seven days. We decrease the number of gestures for the right hand due to lower classification accuracy on the right hand for this user. (Right) Gestures are mapped to robot movements in three different modes. Modes are cycled between either through voice commands (``Hey Robot, Next Mode'') or holding the wrist back gesture on both hands for 0.2 seconds.}
    \label{fig:gestures-and-mappings}
\end{figure*}

\subsection{Data Collection System}
EMG data were collected daily using a cue-based gesture protocol with a brief maximum voluntary contraction (MVC) calibration. The signals were filtered using a 20~Hz high-pass filter and a 60~Hz notch filter, segmented into 80~ms windows with a 40~ms stride, and converted into spatial heat maps of root-mean-square for each forearm. The classifiers were retrained at the discretion of the researcher when online performance was degraded, most often due to a shift in sleeve or changes in the impedance of the skin-electrode. This design prioritized rapid retraining and robustness over time in an in-home setting, with a median retraining time of two times a day over more than four hours. Full details of the data collection and training protocol are provided in the Supplementary \ref{supplementary:data-collection}. 

\subsection{Personalization of Gestures and ML-based Interface}
\label{sec:gesture-personalization}

We collected data from two users with cervical spinal cord injury to evaluate our ML-based interface's gesture classification accuracy, as shown in Figure~\ref{fig:data-collection-two-users}. User 1 is a 39 year-old male (19 years since injury) user with a C5 \ac{cSCI} and an AIS A score, indicating clinically complete motor and sensory quadriplegia. User 2 is a 74 year-old male (7 years since injury) user with a C6 \ac{cSCI} and an AIS B score, indicating clinically complete motor quadriplegia with partial sensory quadriplegia. We plan to release these data as the first open-source EMG dataset from users with quadriplegia upon acceptance of the paper. 

To accommodate individual differences in residual neuromotor activity after cervical spinal cord injury, the gesture-based interface was designed to be personalized to the user. Rather than relying on a fixed set of predefined gestures, we tailored the gesture set to the user's measurable EMG patterns of the forearm, allowing the control interface to adapt to the user’s specific residual motor capabilities. After running a screening study, we selected an RMS heatmap-based preprocessing method with a CNN-based architecture, which we include more details on in Section~\ref{sec:screening-study}. 

\section{Teleoperation System Overview}
\label{sec:teleoperation-module}
This section describes the real-time teleoperation pipeline that connects wearable EMG sensing, gesture classification, user feedback, and robot control in the user’s home.

EMG voltages are acquired using two Intan RHD2164 integrated circuits per arm and streamed to a companion laptop via the Intan recording controller. In total, 256 EMG channels are transmitted over TCP at 4{,}000~Hz to the laptop, where all real-time signal processing and classification are performed.

Incoming EMG signals are filtered using a 60~Hz notch filter and a 20~Hz high-pass filter, then segmented into 80~ms windows. For each window, the root-mean-square magnitudes are computed and arranged as spatial heat maps for each forearm. These heat maps are classified in real time using separate 2D CNN models for the left and right arms, producing a predicted gesture for each hand at each timestep.

Raw predictions are temporally filtered using an exponential moving average of classifier probabilities ($\alpha=0.5$), confidence thresholding at 0.75, and majority voting over 11 consecutive predictions. A gesture command is issued only when at least 6 of the 11 predictions agree, resulting in an effective dwell time of approximately 240–440~ms depending on the consistency of the prediction. Predictions below the confidence threshold are labeled as ``Rest'' to reduce false positives that could trigger unintended robot motion. Robot commands are transmitted at a fixed rate of 10~Hz. 

Commands were rate-limited to 10 Hz and gated by dwell and confidence constraints, which prevented transient misclassifications from producing motion. During active control, unintended motions were rare and typically limited to small movements of the base or wrist that were easily corrected by the user.

\begin{figure}[ht]
  \centering
  \includegraphics[width=0.9\linewidth]{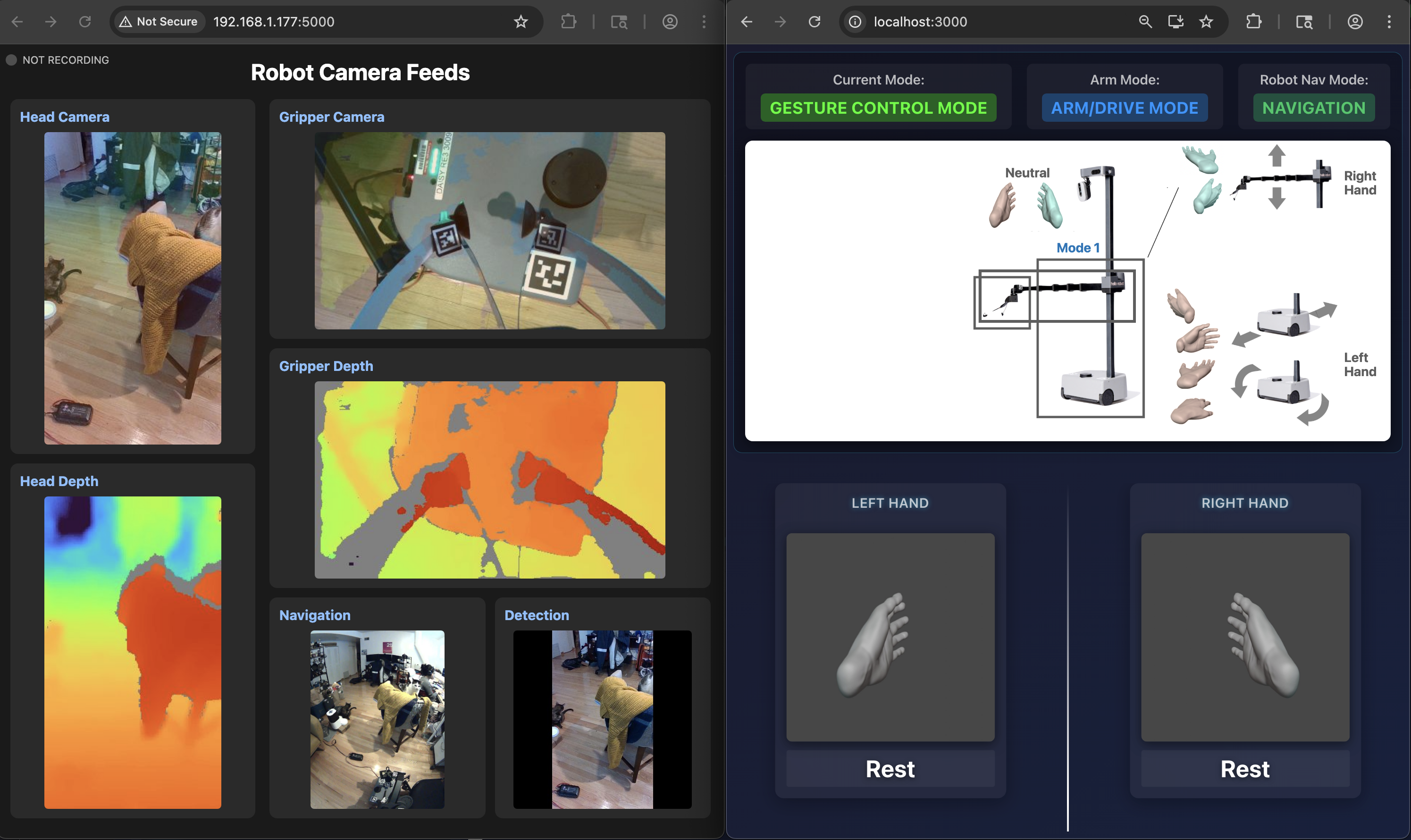}
  \caption{GUI showing camera feeds from the robot, mode mappings, and currently predicted gestures on both hands}
  \label{supplementary:gui}
\end{figure}

The user interacts with the system through a web-based interface, shown in Figure~\ref{supplementary:gui}, which provides real-time feedback on predicted gestures, camera views, and control mappings. This interface is critical for tasks in which the robot operates outside the user’s direct line of sight or at a distance from the user, such as when navigating between rooms or reaching objects in a different room. The camera feeds allow the user to judge depth, alignment, and gripper pose relative to objects, which would otherwise be difficult to infer~\cite{yang2025high}, particularly when aligning the wrist or end effector with a target.

The interface includes a neutral mode in which gesture predictions are displayed without sending commands to the robot, as well as an active gesture mode that can be enabled using a voice command. When gesture mode is active, the interface displays the current mapping between gestures and robot motions for three control modes: (1) Arm and Drive, (2) Arm and Gripper, and (3) Wrist. The mapping between gestures and robot motion across these modes is illustrated in Figure~\ref{fig:gestures-and-mappings}. Gesture commands are transmitted from the companion laptop to the robot using UDP over a local wireless network. The hardware used in this system is illustrated in the Supplementary Figure~\ref{supplementary:hardware}.

\subsection{Mobile Manipulator Robot and Control Mapping}
\label{sec:mobile-manipulator}
We use the Hello Robot Stretch 3 robot with the Dex Wrist attachment, which includes yaw, pitch, and roll wrist movements. The robot has 3 camera feeds, which include two Intel RealSense RGBD cameras, one mounted on the head and one on the wrist, as well as a navigation RGB camera with a larger field of view mounted to the robot head. These three camera feeds, including depth maps, are shown to the user during all tasks using a web interface, as shown in Figure \ref{supplementary:gui}. We also show a \textit{Detection} image camera stream, which shows bounding boxes over objects when detected using a YOLO-E-v26 open-vocabulary perception system, which is described in Section \ref{sec:auto-align}. 

The robot joint speeds are tuned to balance precise manipulation and efficient gross motion. For each joint, a slower speed is used when a command is sustained for three seconds or less, enabling fine positioning, while a higher speed is activated after three seconds to support larger movements. For base translation, arm, and wrist joints, the fast speed is set to twice the slow speed, while for base rotation, it is increased to four times the slow speed to allow for careful alignment during manipulation and efficient turning when navigating. Concretely, the default speeds are listed in the Supplementary Table~\ref{tab:robot_joint_speeds}.

\subsection{Voice Commands}

\label{sec:voice-commands}
We use voice commands to provide the user with a low-effort, high-level interaction channel that allows the user to control system state and query the interface without interrupting gesture-based teleoperation. 

Using voice commands, the user can start and stop gesture mode, which starts and stops gesture commands from being sent to the robot, as shown in the Supplementary Figure \ref{supplementary:voice-switching-modes}. The automatic speech recognition module is implemented using a large-v3 Whisper automatic speech recognition model. Keywords are used to recognize intended mode switches. Additionally, when the user asks questions about the interface, a local LLM implemented using Phi-4 mini answers the questions about the interface. 
Voice responses are generated using Kokoro TTS running on the companion laptop. We show text queries used by the user for auto-alignment trials in the Supplementary Table~\ref{supplementary:tab-text-queries}. We also show a system diagram showing the finite state machine representing the modes and mode transitions using voice commands in the Supplementary Figure~\ref{supplementary:voice-switching-modes}.

\begin{figure}[t!]
    \centering
    \includegraphics[width=0.9\linewidth]{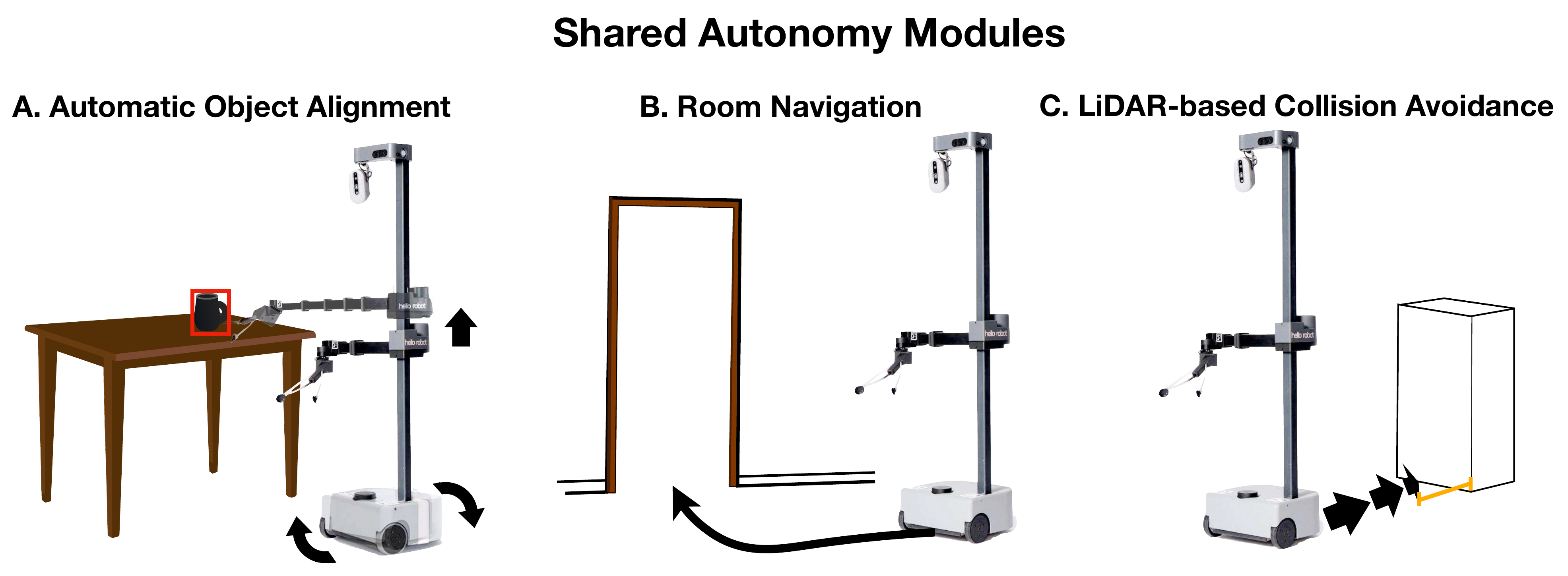}
    \caption{Illustrations of the three shared autonomy modules used: automatic object alignment, room navigation, and LiDAR-based collision avoidance. }
    \label{fig:shared-autonomy}
\end{figure}

\section{Shared Autonomy Modules}
\label{sec:shared-autonomy}
Shared autonomy modules were incorporated to assist with object alignment, room-level navigation, and collision avoidance during in-home teleoperation. These modules provide assistance with coarse spatial alignment and navigation while preserving user authority. An overview of the three shared autonomy modules used in this system is shown in Figure~\ref{fig:shared-autonomy}.

\subsection{Automatic Object Alignment}
\label{sec:auto-align}

We implement an automatic object alignment module inspired by prior work on shared autonomy for assistive manipulation~\cite{padmanabha2024independence}, adapted to our ROS2-based system and updated perception and inverse kinematics pipelines. The module uses an open-vocabulary object detector to identify user-specified objects and assists with coarse alignment of the robot base and arm prior to grasping. 
Object detections are filtered using a confidence threshold of 0.3 and run at 1.5~Hz.

The user activates automatic object alignment via a voice command and specifies the target object verbally (Fig.~\ref{fig:shared-autonomy}A). Unlike a system that uses predefined text queries specified during system design~\cite{padmanabha2024independence}, our approach supports natural language voice queries issued by the user at runtime, allowing object references to adapt dynamically to the needs and environment of the user. Example queries used during the study are shown in the Supplementary Table~\ref{supplementary:tab-text-queries}.

Once an object is detected in the head-mounted RGBD camera, its 3D centroid is computed using known camera intrinsics and extrinsics. An inverse kinematics solver implemented in Pinocchio robotics library is then used to determine a configuration that aligns the robot arm with the object. 

We adopt the Driver Assistance formulation introduced by \citet{padmanabha2024independence} for shared control during object alignment. The Stretch robot is modeled as a nonholonomic mobile manipulator and control is shared between user input and an autonomous alignment controller operating in Cartesian space. 
\begin{equation}
\boldsymbol{u} = \boldsymbol{u}_h + \alpha \boldsymbol{u}_a,
\end{equation}
where $\boldsymbol{u}_h$ is the user’s teleoperation command, $\boldsymbol{u}_a$ is an autonomous alignment command calculated using inverse kinematics toward the detected object and $\alpha \in [0,1]$ modulates the level of assistance based on the confidence of the system. As in previous work, autonomous assistance is constrained to alignment motions and does not override user control of approach or grasp execution.

\subsection{Room Navigation}
  Room-level navigation can be particularly challenging for users when the robot is outside their line of sight or when long-distance base movements are required using a low degree-of-freedom interface.  
  
  Using the Stretch Nav2 package, we implement an automatic room navigation system with shared autonomy. The shared autonomy mode of room navigation is shown in Figure~\ref{fig:shared-autonomy}B. This is activated with voice commands in the interface: the user switches to ``room mode'' and then indicates the destination room (e.g., ``Go to the kitchen.''). The mapping of the home environment is done once by the researcher, who also records the goal coordinates for each room. The two rooms used in this study, the kitchen and the bedroom, are separated by a narrow hallway. For motion planning, the system uses the Stretch Nav2 package, which employs Adaptive Monte Carlo Localization with a pre-built map, Dijkstra's algorithm on a grid-based costmap for global path planning, and a dynamic window approach for local planning.
  
  In each control cycle, the user provides a directional control input $\mathbf{u}_{h,b} = [u_f,\, u_l,\, u_r,\, u_b]^\top$, where $u_f$, $u_l$, $u_r$, $u_b \geq 0$ represent the forward, left turn, right turn and backward commands for the base, respectively. These are generated from the user's gesture interface. Simultaneously, the Nav2 planner produces a planned velocity command $(v_\text{nav},\, \omega_\text{nav})$ consisting of a linear and angular velocity to follow the global path. The output velocity command $(v_\text{out},\, \omega_\text{out})$ sent to the robot is calculated through a combination of user input and planner output. When the user commands backward motion ($u_b > 0$), this completely overrides the planner:
  \begin{equation}
    v_\text{out} = -u_b, \quad \omega_\text{out} = 0.
  \end{equation}
  When the user commands forward and/or turning with $u_b = 0$, the forward component of the planner velocity is gated by the user's forward input: 
  \begin{equation} 
    \alpha =  
    \begin{cases}
      u_f & \text{if } u_f > 0, \\ 
      0.3 \cdot \max(u_l, u_r) & \text{if } u_f = 0 \text{ and } (u_l > 0 \text{ or } u_r > 0), \\ 
      0 & \text{otherwise,}
    \end{cases}
  \end{equation}
  where $\alpha$ is the forward scaling factor. The desired angular velocity of the user is calculated as $\omega_\text{user} = u_l - u_r$, and the scaled planner angular velocity is $\omega_\text{plan} =
  k_\omega \cdot \alpha \cdot \omega_\text{nav}$, where $k_\omega$ is an angular scaling gain. The output velocities are then: 
  \begin{equation} 
    v_\text{out} = k_v \cdot \alpha \cdot v_\text{nav}, 
  \end{equation} 
\begin{equation}
  \omega_{\text{out}} =
  \begin{cases}
    \omega_{\text{user}}, &
    \begin{aligned}
    &\text{if } \operatorname{sgn}(\omega_{\text{user}}) \neq 
    \operatorname{sgn}(\omega_{\text{plan}}) \\
    &\text{and } \omega_{\text{user}} \neq 0
    \end{aligned} \\
    \omega_{\text{plan}} + \omega_{\text{user}}, & \text{otherwise.}
  \end{cases}
\end{equation}
where $k_v$ and $k_\omega$ are linear and angular scaling gains. When the user's turning direction conflicts with the planner's, the user's command takes full precedence, allowing the user to override the planned trajectory when needed. When both agree in direction, the commands are summed, enabling the user to augment the planner's turning. When only the user presses forward ($\omega_\text{user} = 0$), the system follows the planned path. This formulation ensures that the robot only moves along the planned path when the user actively indicates the forward gesture, giving the user continuous control over the progress along the trajectory while retaining the ability to deviate from it.                                                                          

\subsection{LiDAR-based Driving Collision Monitor}
A LiDAR-based collision monitoring module is continuously running on the robot to reduce the risk of base collisions during teleoperation. The collision avoidance behavior during the base motion is illustrated in Figure~\ref{fig:shared-autonomy}C. The module monitors obstacles detected by the 2D LiDAR in front of and behind the robot and progressively scales the base velocity as the robot approaches nearby objects. Obstacles are considered when detected within 0.5~m laterally or 0.3~m in front of or behind the robot. 

Formally, in each cycle, the minimum obstacle distance in the robot's forward path $d_\text{front}$ and backward path $d_\text{back}$ are computed by filtering scan points that fall within a lateral corridor of width $w$ centered on the robot's heading. The velocity is then scaled by: 
\begin{equation} 
\mu = \operatorname{clamp}\!\left(\frac{d - d_\text{offset}}{d_\text{slow}},\; 0,\; 1\right),                                                                  
\end{equation}
where $d$ is the minimum obstacle distance ($d_\text{front}$ or $d_\text{back}$ depending on the direction of travel), $d_\text{offset}$ is the distance from the lidar sensor to the edge of the robot chassis, $d_\text{slow}$ is the threshold of the slowdown distance and $\operatorname{clamp}$ restricts the value to $[0, 1]$. The linear output velocity is then $v_\text{out} \leftarrow \mu \cdot v_\text{out}$. This brings the robot smoothly to a stop as it approaches obstacles.

Arm and wrist interactions were visually monitored by the user through real-time camera feedback, but were not governed by explicit collision prediction or constraint-based control. Safety is further implemented in the study design by the researcher being nearby to stop the robot, the inherent light weight of the robot, as well as the built-in safety mechanisms, detailed in the Supplementary \ref{supplementary:safety}.


\section{Results}
\subsection{Screening Study}
\label{sec:screening-study}
We conducted an initial screening session (\(n=2\)) to characterize the residual EMG activity of the forearm of users and identify gestures that are both separable and reliable for users. User 1 attempted 23 candidate gestures (Supplementary~\ref{supplementary:screening-gestures}) while User 2 attempted 10 candidate gestures (the first 10 gestures listed in Supplementary~\ref{supplementary:screening-gestures}). For each gesture, we computed average spatial activation heatmaps and used cosine similarity to assess separability across gesture pairs. This process enabled the selection of a subset of gestures that was personalized to the residual neuromotor patterns of the user rather than based on a predefined or generic set of gestures. Although we found that more gestures were able to be classified for User 2 (up to all ten), we tailored the latter in-home robot control to User 1, who has a more clinically severe quadriplegic condition due to reduction in motor function of the user's hands and wrists. The criteria we focus on in particular is that the accuracies of the gesture test accuracies over time have to perform above 85\%, which we find generally enables consistent classification when using the gesture recognition system for effective robot control. 

\begin{table}[ht]
  \centering
  \caption{Mean validation accuracy (\%) by feature and architecture, averaged across learning rates.}
  \label{tab:sweep_val_mean}
  \begin{tabular}{lccc}
  \toprule
  Feature & MLP & CNN & TDS \\
  \midrule
  MPF  & $88.4 \pm 3.3$ (10) & $88.5 \pm 1.3$ (12) & $89.1 \pm 2.7$ (12) \\
  RMS  & $85.5 \pm 6.8$ (5)  & $\mathbf{92.4 \pm 0.6}$ (8) & $89.9 \pm 2.7$ (6) \\
  Raw  & $82.9 \pm 4.3$ (8)  & $88.5 \pm 0.8$ (7)  & $88.5 \pm 2.7$ (7) \\
  CWT  & $90.2 \pm 1.1$ (9)  & $90.6 \pm 1.1$ (2)  & $89.2 \pm 2.7$ (5) \\
  STFT & $86.6 \pm 1.3$ (9)  & $87.1 \pm 2.0$ (10) & $88.9 \pm 1.7$ (8) \\
  \bottomrule
  \end{tabular}
  \\[2pt]\footnotesize Mean $\pm$ std val/accuracy (\%) over all runs per cell (learning rates and seeds pooled); $n$ in parentheses. Bold = best.
\end{table}

\begin{table*}[ht]
    \centering
    \caption{5-fold CV accuracy (\%) for five class classification (mean, std across 4 hands for 2 participants) and estimated
    full end-to-end wall clock. Abbreviations include lin for linear probing, and ft for fine-tuning, where ResNet18 and TinyViT are pretrained on ImageNet. TDS and MetaCNN-LSTM are designed to be run with channel \(\times\) temporal features, meaning that they were only run with the Raw and MUAP feature extraction methods.}
    \label{tab:five-fold-cv}

    \begin{tabular}{l|ccccccccc}
      \toprule
      \textbf{Features}
      & MLP
      & CNN
      & \makecell{Transformer}
      & TDS
      & \makecell{MetaCNN-\\LSTM}
      & \makecell{ResNet18\\lin}
      & \makecell{ResNet18\\ft}
      & \makecell{TinyViT\\lin}
      & \makecell{TinyViT\\ft} \\
      \midrule

      \textbf{RMS}
      & \makecell{79.8 \(\pm\) 18}
      & \makecell{\textbf{81.5} \(\pm\) 18}
      & \makecell{68.2 \(\pm\) 10}
      & --
      & --
      & \makecell{75.1 \(\pm\) 16}
      & \makecell{58.2 \(\pm\) 17}
      & \makecell{74.6 \(\pm\) 15}
      & \makecell{34.4 \(\pm\) 8} \\

      \addlinespace

      \textbf{Raw}
      & \makecell{46.8 \(\pm\) 10}
      & \makecell{73.9 \(\pm\) 24}
      & \makecell{63.1 \(\pm\) 11}
      & \makecell{80.6 \(\pm\) 19}
      & \makecell{79.9 \(\pm\) 19}
      & \makecell{75.0 \(\pm\) 16}
      & \makecell{78.2 \(\pm\) 19}
      & \makecell{74.8 \(\pm\) 16}
      & \makecell{69.7 \(\pm\) 21} \\
  
      \addlinespace

      \textbf{CWT}
      & \makecell{81.4 \(\pm\) 20}
      & \makecell{\textbf{82.5} \(\pm\) 18}
      & \makecell{69.5 \(\pm\) 22} 
      & --                                                                                                                                                                                                                                                                                                          
      & --
      & \makecell{69.6 \(\pm\) 13} 
      & \makecell{64.1 \(\pm\) 22}
      & \makecell{68.3 \(\pm\) 11}
      & \makecell{25.6 \(\pm\) 8} \\
  
      \addlinespace

      \textbf{MUAP}                                                                                                                                                                                                                                                                                                 
      & \makecell{63.6 \(\pm\) 17}
      & \makecell{62.9 \(\pm\) 16}
      & \makecell{42.2 \(\pm\) 13} 
      & \makecell{62.4 \(\pm\) 17}                                                                                                                                                                                                                                                                                  
      & \makecell{61.8 \(\pm\) 17}
      & \makecell{59.8 \(\pm\) 17} 
      & \makecell{61.8 \(\pm\) 18}
      & \makecell{58.6 \(\pm\) 16}
      & \makecell{54.0 \(\pm\) 16} \\

      \bottomrule
    \end{tabular}
  \end{table*}

Based on data from the screening study, the five selected gestures for the in-home study phase based on User 1's data were: (1) rest, (2) wrist forward, (3) wrist back, (4) wrist supination, and (5) wrist pronation. The EMG data from each hand are represented as a spatial root-mean-squared (RMS) heatmap and classified using a lightweight \ac{CNN}. This feature preprocessing method and neural network architecture were selected from an evaluation of several architectures and preprocessing types using a random hyperparameter search summarized in Table~\ref{tab:sweep_val_mean} using data collected during the screening study. This evaluation was performed for User 1 on the user's right hand. The highest performing combination was performed on a left-out test set of the last 6 gesture trials for each gesture (of 61 gesture trials), which produced a test set accuracy performance of \(92.3\% \pm 0.4 \) (n=3), close to the validation set performance of \(92.4\% \pm 0.6\). We define a single gesture trial as when a user attempts to perform a gesture, ending when the user relaxes afterwards.

To validate our preprocessing and architecture selection, we also compare our preprocessing and architecture selection on a more diverse dataset with a more rigorous evaluation: across both hands of both users with \ac{cSCI} using five-fold cross validation with 15 gesture trials from each hand. Table~\ref{tab:five-fold-cv} reports these results, including architectures and preprocessing techniques adapted from those presented by \citet{kaifosh2025generic} (MetaCNN-LSTM) and \citet{yang2025intuitive} (motor unit action potential (MUAP) preprocessing). Details on preprocessing and architectures are provided in Supplementary Sections~\ref{supplementary:details-on-preprocessing} and ~\ref{supplementary:details-on-architectures}, respectively. An additional breakdown of individual user performance is presented in the Supplementary Table~\ref{tab:per-hand-cv}, where User 1 consistently has a lower classification accuracy for the five gestures than User 2 due to the higher clinical severity of paralysis in the upper body from \ac{cSCI}. 

Although a continuous wavelet transform (CWT) method with the CNN architecture performed slightly better, due to the training of the model taking 4 times longer, the RMS heatmap preprocessing method was chosen instead for the 12-day study. The selected preprocessing method and architecture enables rapid retraining following new data collection sessions, which is essential for real-time use during a multi-day in-home study. The CNN consists of two 2D convolutional layers with $3 \times 3$ kernels and 128 filters, followed by batch normalization and ReLU activations. The resulting features are flattened and passed through a fully connected 128-node layer for classification. 

\subsection{Twelve-Day In-Home Study}
\label{sec:in-home-study}
We conducted a follow-up twelve-day in-home study with User 1, a 39 year-old male (19 years since injury) user with a C5 \ac{cSCI} and an AIS A score, indicating clinically complete motor and sensory quadriplegia. The user had no previous experience controlling a mobile manipulator and had not previously used a robot in their home. For clarity, we rank study days consecutively from Day~1 to Day~12. Days~1--5 correspond to the exploratory phase, while Days~6--12 correspond to the finalized study phase (referred to as Week~2). Shared autonomy modules were introduced incrementally during the finalized study phase, with automatic object alignment enabled on Day 9 (Week 2, Day 4) and room navigation enabled on Day 11 (Week 2, Day 6).

\subsubsection{Five-Day Exploratory Phase}
During the five-day exploratory phase with User 1, we evaluated whether a symmetric five-gesture vocabulary on both hands was feasible for sustained robot control. Although the left-hand classification performance remained high (mean test precision $90.9\% \pm 5.3\%$), the right-hand performance was substantially lower (mean test precision $76.6\% \pm 4.4\%$), reflecting the asymmetries in the residual neuromotor activity of the user. To improve reliability during real-time control, we personalized the interface by reducing the right-hand gesture set to three gestures for the finalized study phase. This reduction improved reliability during real-time control by removing gestures that were inconsistently classified, at the cost of reducing the number of available commands. The final sets of gestures used for robot control in each hand are illustrated on the left of Figure~\ref{fig:gestures-and-mappings}. This personalized configuration remained stable throughout the finalized study phase and supported reliable daily use without requiring precise fixed anatomical placement or excessive calibration sessions from additional data collection.

\begin{figure}[ht]
    \centering
    \includegraphics[width=1.0\linewidth]{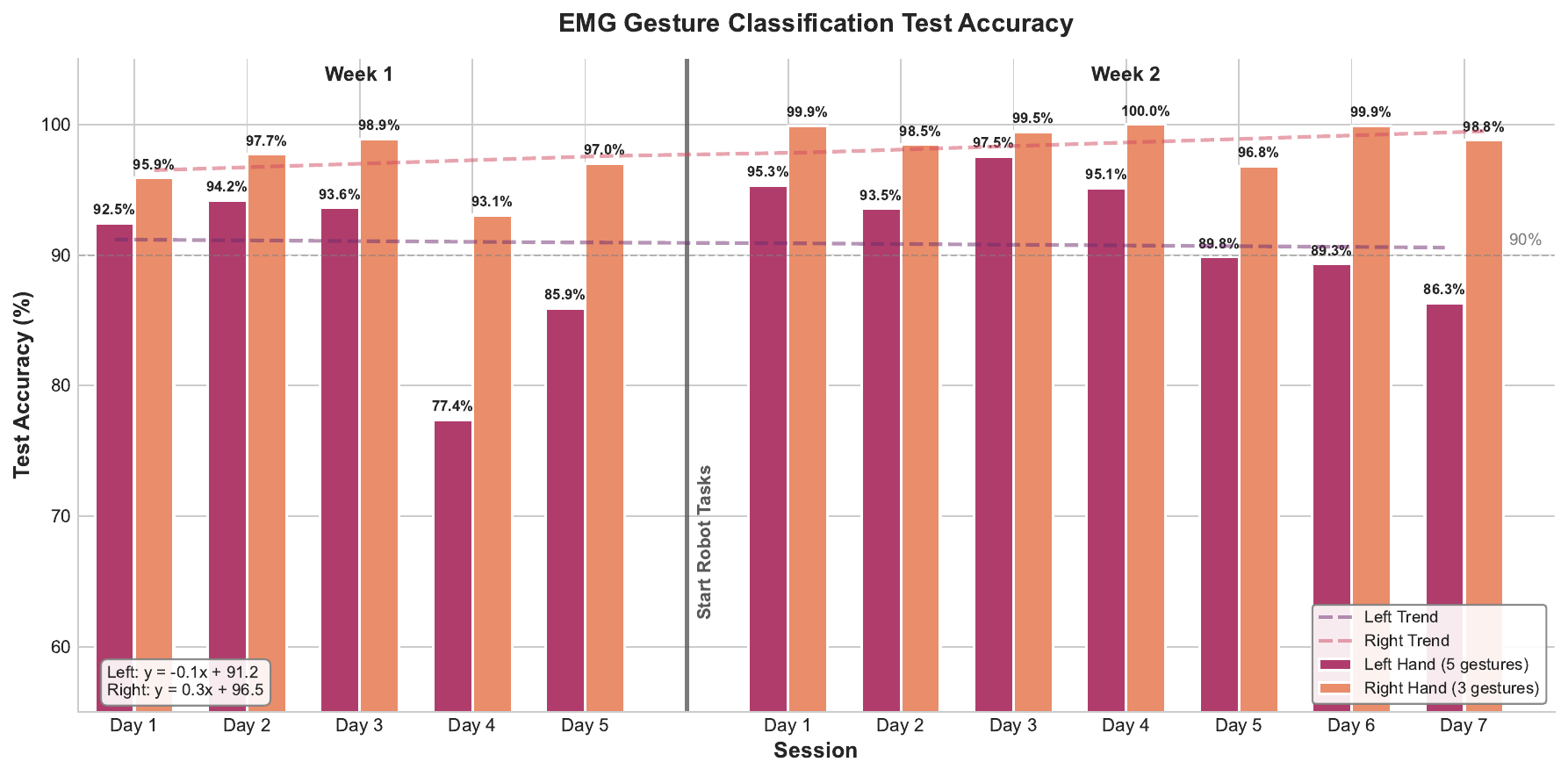}
    \caption{Test accuracies per day for week 1 and week 2. Test accuracies involve accuracies performed on the last gesture trials performed with each gesture for the session.}
    \label{fig:test-accuracy}
\end{figure}

\begin{figure*}[t]
  \centering
  \includegraphics[width=0.95\linewidth]{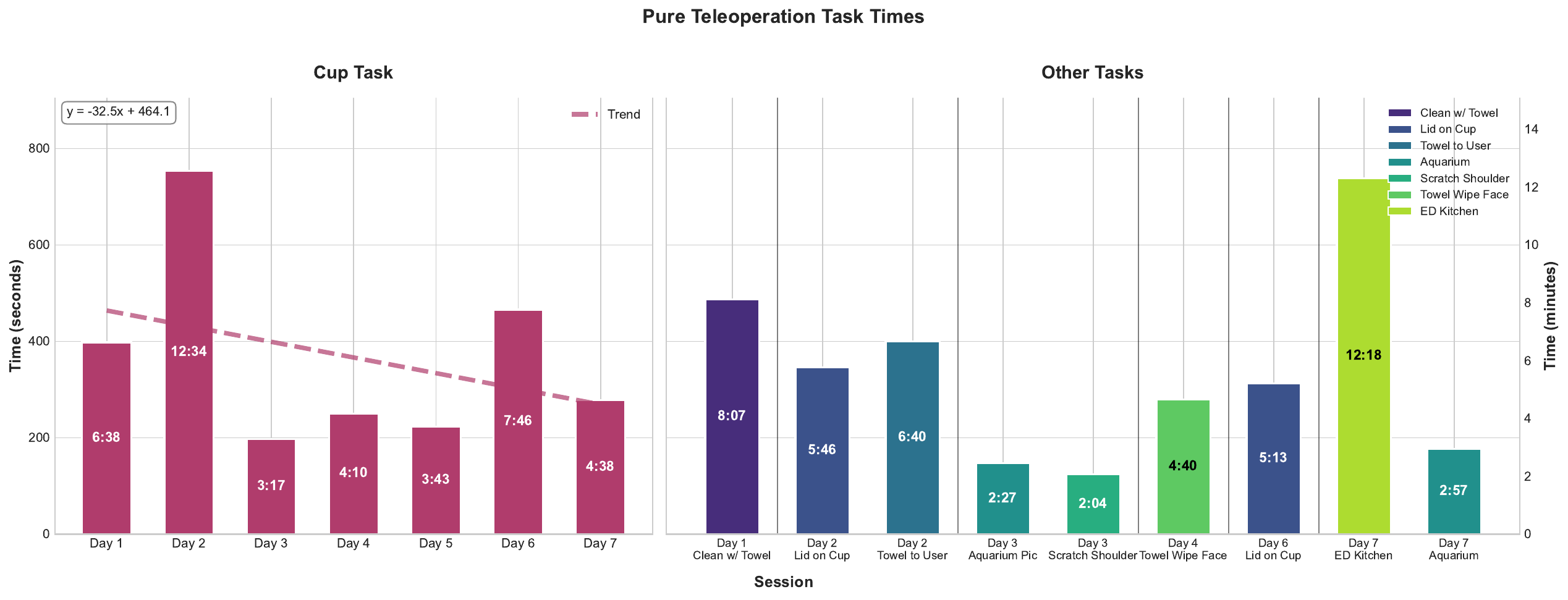}
  \small
    \captionof{figure}{Task times for all tasks performed over the second week. (Left) Task times for cup task performed over the second week. This task is repeated each day. A least-squares trend line is shown, indicating a decrease of approximately 33 seconds per day. (Right) Task times for other tasks performed over the second week.}
    \label{fig:teleop-times}
\end{figure*}

We report daily classification performance for the five-gesture configuration on both hands during Days~2--5 in the Supplementary Figure~\ref{supplementary:test-accuracy-week1-5-gestures} with confusion matrices in the Supplementary Figure~\ref{supplementary:confusion-matrices}. The classification performance for the final configuration on the first five days is shown on the left in Figure~\ref{fig:test-accuracy}.  We note that test classification accuracy varied each day, in part due to the user's condition for each day. For example, on Day 4 of Week 1, the user mentioned that they had not slept the night before and appeared to have missed multiple gesture cues during EMG data collection.


From Day~2 through Day~4, we evaluated the EMG classifiers by having the user perform a subset of robot control tasks, described in Supplementary~\ref{supplementary:sec-practice-tasks}. Performance during the practice tasks is shown in Supplementary Figure~\ref{supplementary:practice-tasks}. 

Insights from the practice phase informed refinements to the teleoperation interface. In particular, we added additional camera feedback to the visual interface for the GUI shown in Figure~\ref{supplementary:gui}, including an in-depth wrist camera view and a wide-angle navigation camera, after the user reported difficulty judging gripper distance and spatial relationships between the robot and surrounding objects. These additions were intended to help the user judge depth, gripper distance, and robot alignment relative to objects, and to reduce the likelihood of wrist collisions with the environment.

\subsubsection{Finalized Seven-Day Robot Control Study}

The subsequent seven-day study phase used a standardized teleoperation setup with a fixed number of camera feeds throughout all sessions. In parallel, the gesture classification system was personalized to use only three gestures on the user’s right hand. This reduction enabled a more reliable real-time classifier and reduced observed robot control errors due to misclassifications.

\subsubsubsection{Gesture Classification Accuracy}
The daily test accuracies for both hands in the final study phase is shown on the left of Figure~\ref{fig:test-accuracy}. Confusion matrices showing lower classification accuracies for wrist forward and wrist pronation are shown in the Supplementary Figure~\ref{supplementary:confusion-matrices}. We note an average test accuracy and standard deviation of 90.9\% ($\pm 5.3\%$) for the left hand and 98.0\% ($\pm 2.0\%$) for the right hand. These right-hand accuracies correspond to the reduced three-gesture configuration used during the finalized study phase.

This is comparable to test accuracies seen using classification of motor units in \citet{yang2025non}, where there is an average of 82\% test accuracy for 8-gesture classification for 8 subjects with \ac{cSCI} when testing with 62.5ms windows of EMG. We note that all subjects tested in \citet{yang2025non} have less severe or equal \ac{cSCI} than our User 1 according to AIS scores. By plotting trend lines using least-squares, we see that there seem to be small slopes, showing an insignificant change in test classification accuracy from the first day to the last day of the study.
These results suggest that for assistive teleoperation, robustness of temporal filtering and user adaptation may be more important than marginal gains in offline classification accuracy.



\subsubsubsection{Benefits of Real-Time EMG Classification Modules}
To characterize the benefits of the real-time dwell and confidence modules, we found that simple majority voting and exponential moving average (EMA) increase classification accuracy, as seen in Table \ref{tab:smoothing_avg_std} evaluated on data for 12 days from User 1.
\begin{table}[ht]
\centering
\scriptsize
\caption{Average classification accuracy across days under smoothing.}
\begin{tabular}{lcccc}
\toprule
 & Raw & EMA Only & Maj. Only & EMA + Maj. \\
\midrule
Left (5 gest.)  & 90.9 \(\pm\) 5.3\% & 90.3 \(\pm\) 8.1\% & 91.1 \(\pm\) 8.1\% & \textbf{92.3 \(\pm\) 7.6\%} \\
Right (3 gest.) & 98.0 \(\pm\) 2.0\% & 97.4 \(\pm\) 2.6\% & 97.9 \(\pm\) 2.5\% & \textbf{98.5 \(\pm\) 2.4\%} \\
\bottomrule
\end{tabular}
\label{tab:smoothing_avg_std}
\end{table}
Additionally, we found that the false positive rates (FPR) of the online classifier each day for the second week of User 1 (where we collected additional data having the user follow cued gestures using a deployed classifier) were reduced, as shown in Table \ref{tab:fpr_avg_std}, using EMA + majority voting. We follow the same procedure for a cue-based online classification evaluation as in \citet{yang2025high}. 
\begin{table}[ht]
\caption{Average false positive rate across days for online classification evaluation in second week.}
\centering
\scriptsize
\begin{tabular}{lcccc}
\toprule
 & Raw & EMA Only & Maj. Only & EMA + Maj. \\
\midrule
Left (5 gest.)  & 4.2 \(\pm\) 2.9\% & 0.4 \(\pm\) 0.6\% & 0.3 \(\pm\) 0.6\% & \textbf{0.1 \(\pm\) 0.2\%} \\
Right (3 gest.) & 8.3 \(\pm\) 11.3\% & 0.6 \(\pm\) 0.7\% & 0.6 \(\pm\) 0.8\% & \textbf{0.5 \(\pm\) 0.7\%} \\
\bottomrule
\end{tabular}
\label{tab:fpr_avg_std}
\end{table}

\subsubsubsection{Tasks}
We evaluated the robot control system on a set of tasks that span \ac{ADLs}, instrumental ADLs, and a personalized leisure task selected by the user. Tasks included drinking from a cup, placing a lid on a cup, wiping surfaces with a towel, retrieving an energy drink from another room, and assisting with self-contact tasks such as wiping the face or scratching the shoulder. Pictures of representative tasks performed during the finalized study phase are shown in Figure~\ref{fig:system-overview}, while verbal descriptions are provided in Supplementary~\ref{supplementary:task-setup}. To assess learning over time, the cup-drinking task was repeated daily during the finalized phase of the study. In addition, a personalized task was included that involves taking photos of the user's aquarium to reflect user-specific priorities beyond basic ADLs. 
Additional details on task selection are provided in Supplementary~\ref{supplementary:task-selection}.



\subsubsubsection{Task Times}
During the 12-day study, the user performed 11 different tasks (8 during the finalized seven-day phase), with task completion times ranging from 124--754~s (Fig.~\ref{fig:teleop-times}). Although not directly comparable due to differences in task design and environment, these task times are of the same order as prior HDEMG mobile manipulation results in the lab with able users (mean 3 to 5~min)~\cite{yang2025high}, as well as previous home deployments with users with quadriplegia (151--990~s)~\cite{padmanabha2024independence}. To evaluate the effects of training over time, we focus on the drink task, which involves retrieving a cup from a table, placing the straw near the user’s mouth, and the user drinking. This task was repeated daily during the finalized study phase. Unless otherwise noted, all reported task-time, usability, and workload results refer to the finalized seven-day study phase.

Using least-squares trend lines, the task completion time for the cup task decreased by approximately 33~s/day under pure teleoperation (Fig.~\ref{fig:teleop-times}). 
Although repeated practice alone substantially improved task efficiency under pure teleoperation, shared autonomy was designed to address complementary challenges that persist even with training, including alignment variability, depth perception at distance, and navigation across rooms.

\begin{figure}[ht]
    \centering
    \includegraphics[width=0.7\linewidth]{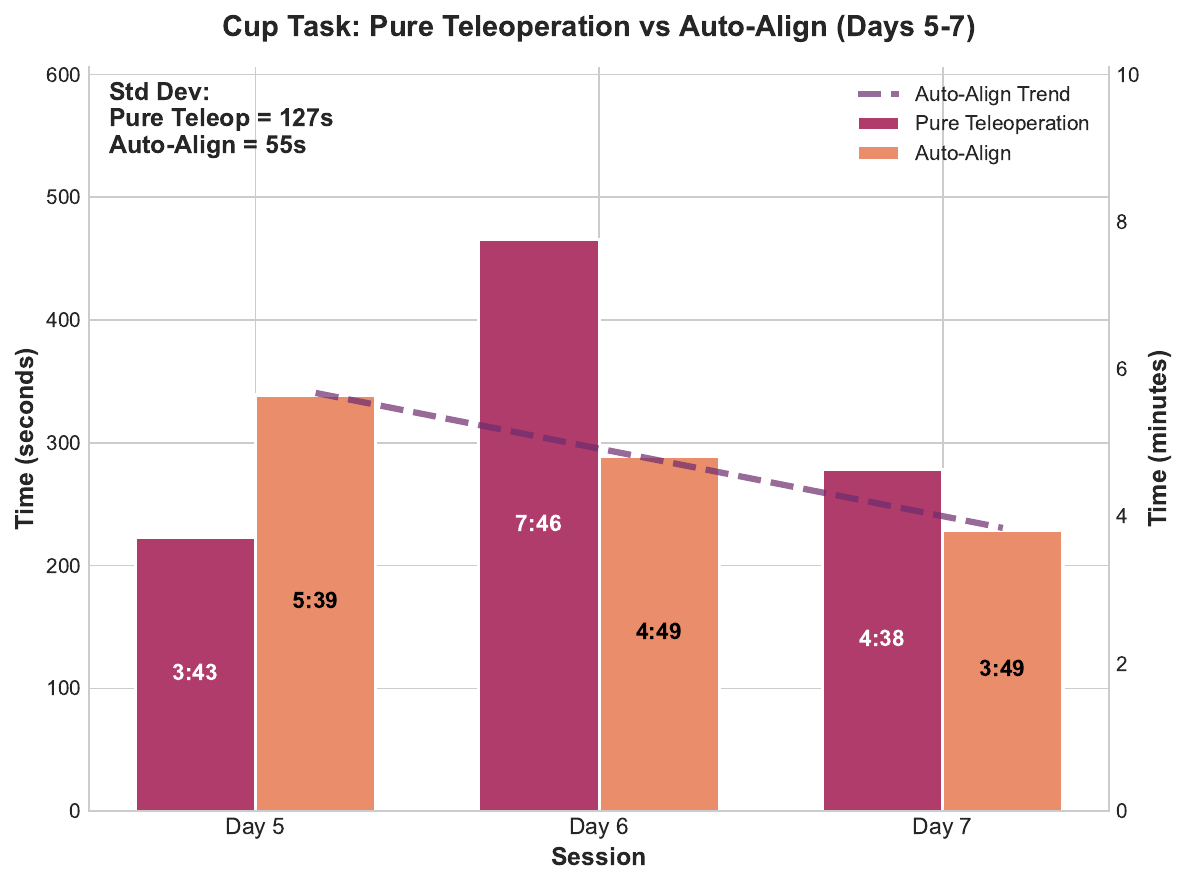}
    \caption{Task times comparing pure teleoperation with auto-align. Auto-alignment shows a visually consistent decrease in task time, while pure teleoperation exhibits higher variability.}
    \label{fig:teleop-vs-shared-last-3}
\end{figure}

\subsubsubsection{Effects of Shared Autonomy}
When auto-alignment was enabled for the cup task, task completion time decreased by approximately 55~s/day based on least-squares trend lines (Fig.~\ref{fig:teleop-vs-shared-last-3}). 

Beyond the mean task time, shared autonomy substantially reduced trial-to-trial variability. As shown in Fig.~\ref{fig:teleop-vs-shared-last-3}, the standard deviation of task completion time under pure teleoperation was $2.3\times$ higher than with auto-alignment. This suggests that manual teleoperation was more sensitive to factors such as fine alignment accuracy, depth perception, and momentary control errors, whereas auto-alignment provided more repeatable alignment and more predictable execution across sessions.

The task-time differences between pure teleoperation and shared autonomy were task-dependent (Supplementary Fig.~\ref{supplementary:task-time-comparisons-all-shared-autonomy}). For tasks performed with close line-of-sight and continuous camera feedback, manual teleoperation sometimes matched the efficiency of shared autonomy, as the user could directly correct alignment using visual feedback without relying on perception-driven assistance.

Deployment in a real home revealed open-vocabulary perception failure modes that are rarely observed in laboratory settings. For example, in the ``lid on cup'' task, the perception system initially misidentified the cup as the lid due to its bicolor appearance and the narrow horizontal field of view of the head-mounted RGBD camera, which delayed the lid entering view until additional base rotation. This misdetection in part increased the task time by 1.25~minutes compared to pure teleoperation for that trial. These failures occurred during normal use with the user’s own household objects rather than due to user error.

Over successive sessions, the user adapted to these constraints by positioning objects within the camera field of view and issuing more specific language queries (e.g., ``cup with lid'' instead of ``cup,'' ``energy drink'' instead of ``can''), leading to improved shared-autonomy performance, as presented in the Supplementary Table~\ref{supplementary:tab-text-queries}. This highlights an important aspect of in-the-wild deployment: effective use of perception-driven assistance depends not only on model capability but also on user familiarity with system behavior.

\begin{table}[ht]
\centering
\caption{Energy Drink Kitchen task times comparing pure teleoperation and align + room mode.}
\label{tab:align-room-vs-teleoperation}
\begin{tabular}{lcc}
\toprule
Condition & Avg. Time (s) & Trials ($n$) \\
\midrule
Pure Teleoperation & 738 & 1 \\
Align + Room Mode  & 517 $\pm$ 62 & 2 \\
\bottomrule
\end{tabular}
\end{table}

Shared autonomy yielded its largest time benefits for navigation-heavy tasks requiring multi-room traversal and alignment with distant objects. For the Energy Drink Kitchen task, the combination of room navigation and auto-alignment reduced task completion time by an average of 30.0\% on Days~6--7, when these modules were enabled (Tab.~\ref{tab:align-room-vs-teleoperation}). Taken together, these results indicate that shared autonomy improves consistency and supports navigation under limited-bandwidth control, with benefits that become more pronounced as users learn to leverage its capabilities in tasks with reduced user perception and at larger spatial scales.

\subsubsubsection{Subjective Results}
From a deployment perspective, donning the HDEMG sleeves required only minor skin preparation and typically took several minutes. Recalibration was performed opportunistically when online performance decreased (median two times per day), most often due to sleeve shift. No safety-critical collisions occurred. 

\begin{figure}[ht]
    \centering
    \includegraphics[width=0.85\linewidth]{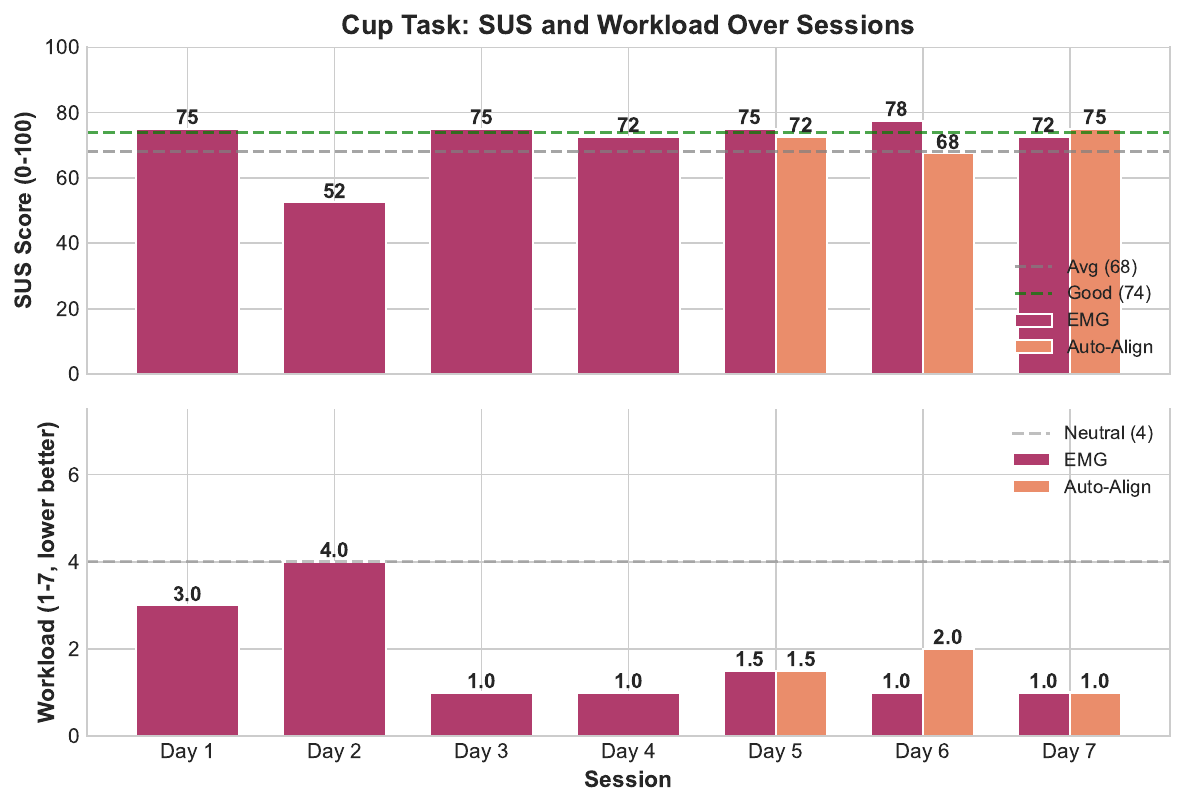}
    \caption{(Top) Week 2 \ac{SUS} of cup task comparing teleoperation vs shared mode. Reference lines correspond to standard SUS interpretive thresholds. (Bottom) 7-point Likert-style median NASA-TLX workload score of cup task for teleoperation vs shared mode.}
    \label{fig:sus-and-workload}
\end{figure}

We evaluated usability and workload using the System Usability Scale (SUS) and a modified 7-point NASA-TLX survey, respectively (Fig.~\ref{fig:sus-and-workload}). The SUS scores remained above the average usability threshold (68) on all days except one and exceeded the good usability threshold on most days~\citep{bangor2008empirical}. Both SUS and NASA-TLX exhibited favorable trends over repeated use, with increasing perceived usability and decreasing workload across days (Figs.~10–11). By the final day of study, shared autonomy achieved shorter task times with comparable workload and higher usability than pure teleoperation, indicating a reduced burden without added subjective cost.

\section{Limitations and Future Work}
\label{sec:limitations}

Daily classifier training required approximately 30 minutes of wall-clock time due to conservative rest periods, despite using as little as 10 minutes of EMG data per session. Because classifiers were trained on limited daily data, they did not explicitly model longer-term nonstationarity arising from sleeve shift or changes in skin--electrode impedance. Although we performed preliminary studies for pretraining from other days and finetuning from the same day for the twelve days of data in the Supplementary Table~\ref{tab:xday_lodo_12day}, we found that fine-tuning performance is mixed, with decreasing fine-tuning performance for the right hand. By releasing our dataset for both users as the first EMG dataset with users with quadriplegia, we hope that additional few-shot adaptation research can be performed with this particular user population to significantly improve performance with recent deep learning techniques by researchers, who generally have weaker neuromotor signals compared to able-bodied users. In practice in our work, performance degradation was mitigated through brief, opportunistic retraining for a subset of gestures. 

Shared autonomy features were introduced incrementally and evaluated over a limited number of days, and results should therefore be interpreted as evidence of feasibility rather than as a comprehensive component-level evaluation. The robot control study work is a single-user in-home case study intended to surface system-level challenges that may not appear in short laboratory studies. The observed trend in task-time differs from prior evaluations of alignment assistance conducted without continuous camera feedback~\citep{padmanabha2024independence}, where manual alignment was substantially more difficult. In our system, persistent visual feedback enabled efficient manual alignment for nearby objects, reducing the marginal time benefit of automation for manipulation-dominated tasks. However, shared autonomy remained beneficial for navigation-heavy tasks involving longer distances and spatial uncertainty, such as multi-room retrieval. Broader validation will require multi-user deployments, comparisons against alternative assistive interfaces, and more systematic evaluation of robustness and voice interaction reliability.

\section{Conclusion}
\label{sec:conclusion}

In this work, we demonstrate the feasibility of a wearable bimanual neuromotor interface to enable real-time control of a mobile manipulator by a user with quadriplegia. Using a fabric-based sleeve incorporating a flexible printed circuit board and high-density EMG sensing, the system captures residual neuromotor signals to support gesture-based robot control during daily use. The sleeve was worn comfortably for extended periods each day and could be activated and deactivated using voice commands, allowing the user to seamlessly transition between robot interaction and rest.

We also present a complete teleoperation system that combines EMG-based gesture control with shared autonomy to support mobile manipulation in home over multiple days. In a 12-day deployment, the user successfully completed a variety of household and instrumental activities of daily living, with repeated task performance that showed reduced completion times, decreased perceived workload, and increased perceived usability. Together, these results suggest that wearable neuromotor interfaces, when combined with shared autonomy, can provide a practical and user-driven approach to physically assistive mobile manipulation in real home environments. 



\bibliographystyle{plainnat}
\bibliography{references}






\renewcommand{\thesection}{S\arabic{section}}
\setcounter{section}{0}

\setcounter{figure}{0}
\renewcommand{\thefigure}{S\arabic{figure}}

\setcounter{table}{0}
\renewcommand{\thetable}{S\arabic{table}}

\section{Supplementary Material}

\subsection{Data Collection Details}
\label{supplementary:data-collection}
During data collection, the user performs gestures using a cue-based protocol. Each trial begins with a visual cue indicating the next gesture for 3 seconds, followed by a 5-second hold period during which the user attempts to perform the gesture. To account for variable reaction times, we use the final 4 seconds of each hold period as training data. At the beginning of each session, the user performs each gesture once at maximum voluntary contraction (MVC) to initialize an RMS-based calibration.

EMG signals are processed using a 20~Hz high-pass filter and a 60~Hz notch filter to remove motion artifacts and noise from the power line. We then compute the RMS value for each channel. For real-time feedback, we use the 90th percentile of RMS values across the 128 channels, which provides a robust measure of activation for gestures that may be spatially localized due to residual motor unit activity. This feedback is displayed to the user during data collection to help regulate the effort of the user.

To encourage consistent signal quality, data collection sessions alternate between target effort ranges of 15--30\% and 20--40\% of the MVC RMS. Each session consists of five sets, with three repetitions of each gesture per set. Depending on the day, the user completes between two and six data collection sessions. When online classification performance degrades during robot control, typically due to sleeve shift or changes in skin impedance, additional data collection sessions are performed at the researcher’s discretion, collecting data from gestures with the lowest online accuracy.

For classifier evaluation, we reserve the final 10\% of gesture trials for each gesture as a held-out test set, enabling evaluation of generalization across time and across repeated gesture attempts. The remaining data is shuffled and split into training and validation sets using a 75/25 split. To accommodate daily time constraints during the study, the classifiers are trained for three epochs. 
Samples are generated from splitting data into 80 ms windows with a 40 ms stride length.

\subsection{Screening Gestures}
We list all the gestures used for the screening. The five gestures we use in the robot control study are in bold:
\label{supplementary:screening-gestures}
\begin{itemize}
    \item Wrist adduction,
    \item Wrist abduction,
    \item \textbf{Wrist flexion},
    \item \textbf{Wrist extension},
    \item Finger extension,
    \item Power grip,
    \item \textbf{Rest},
    \item \textbf{Wrist pronation}, 
    \item \textbf{Wrist supination},
    \item Tripod grip,
    \item Thumb-index-middle extension,
    \item Thumb-index extension,
    \item Thumb flexion,
    \item Thumb extension,
    \item Ring flexion,
    \item Ring extension,
    \item Pinky flexion,
    \item Pinky extension,
    \item Middle flexion,
    \item Middle extension,
    \item Index-middle extension,
    \item Index flexion, and
    \item Index extension.
\end{itemize}

\subsection{Graphical User Interface and Voice Commands}
\label{supplementary:voice-commands}

We present examples of text queries used by the user in Table~\ref{supplementary:tab-text-queries}.

\begin{table}[H]
\centering
\caption{Text Queries Used by User for Auto-Alignment Tasks}
\label{supplementary:tab-text-queries}
\begin{tabular}{cc}
\toprule
Task                  & Text Queries \\ \midrule
Drink from Cup           & ``cup'', ``cup with lid'' \\
Energy Drink from Kitchen     & ``can'', ``energy drink''\\
Lid on Cup & ``lid'' \\ 
\bottomrule
\end{tabular}
\end{table}

\subsection{Robot Built-In Safety Mechanisms}
\label{supplementary:safety}
The robot has a built-in physical run-stop button on the robot head, allowing the researcher or a caregiver to step in to stop the robot motors from moving. Additionally, the built-in motor current limits of the Stretch robot also provided a low-level safety mechanism by stopping joint movement when excessive resistance was detected, which are conservative enough to have low physical risks for the user. For example, in this study, the robot arm and wrist had made contact with the user's torso as a teleoperation error without causing injury.

\subsection{Task Selection}
\label{supplementary:task-selection}
We selected a range of tasks to reflect both commonly identified activities of daily living (ADLs) for people with paralysis and tasks that were personally meaningful to our primary end user. Previous qualitative work examining the needs and expectations of people with quadriplegia and current assistive robotic arm users highlights drinking, grasping and transporting objects, door manipulation, and self-contact tasks as among the most frequently desired capabilities for assistive robotic systems~\cite{styler2025qualitative,hutmacher2025identification}. In particular, these studies emphasize the importance of unilateral reach-and-grasp tasks, manipulation of everyday objects, and interaction with the environment as the main contributors to perceived independence and utility of robotic arms.

Consequently, our ADL task set includes tasks such as drinking from a cup, placing a lid on a cup, wiping surfaces with a towel, and retrieving an energy drink from another room, which align with object manipulation and mobility-related activities described by participants in previous qualitative interviews \citep{styler2025qualitative}. Tasks involving contact with the user’s body, such as scratching the shoulder and wiping the face, reflect additional needs identified in these studies, where users reported difficulty performing self-care actions without external assistance \citep{hutmacher2025identification}. The environmental interaction tasks included in the practice set, such as opening a door and adjusting curtains, similarly mirror common examples of desired robotic assistance discussed by current users of the robotic arm \citep{styler2025qualitative}.

In addition to these ADLs, we incorporate a personalized leisure task, taking photos of the user’s aquarium, which the user explicitly suggested as a meaningful activity they would like to perform independently. Prior work notes that beyond basic ADLs, users value the ability to engage in individualized and context-specific activities that contribute to quality of life and personal identity \citep{hutmacher2025identification}. Because the user lacks finger and wrist flexion control, independently operating a tablet camera is challenging. In our system, the Stretch robot holds the user’s iPad Mini at the appropriate height and orientation, while the user issues voice commands to trigger the camera, enabling completion of this task.

\subsection{Task Setup and Description} 
\label{supplementary:task-setup}
\begin{enumerate}
    \item Drink from Cup \\
    $\bullet$ Setup: A cup of water with a straw is placed on a table to the left of the wheelchair user. \\
    $\bullet$ Description: The user has to drive the robot towards the cup, grab the cup, and then bring the cup to their face. Because the height of the robot is lower than the user's face in his wheelchair, the robot needs to angle up the wrist and roll the wrist to locate the end of the straw around the user's mouth. The task is considered finished when the user drinks from the cup. 
    \item Lid on Cup \\
    $\bullet$ Setup: A cup and a lid is placed on a table next to the user. \\
    $\bullet$ Description: The user has to pick up the lid that is flat on the table, then place the lid on the cup. The task is considered finished when the lid is placed on the cup and pressed onto the cup.
    \item Scratch Shoulder\\
    $\bullet$ Setup: The robot is placed to the left of the user. \\
    $\bullet$ Description: The user has to drive the robot to the user and place pressure with the gripper on the user's shoulder. The task is considered finished when the user has made contact with the robot and rubbed their shoulder with the towel. 
    \item Towel Wipe Table\\
    $\bullet$ Setup: The robot holds a small black towel in its gripper. Five pieces of red masking tape are placed on the table within a small square.\\
    $\bullet$ Description: The user has to remove the five pieces of red tape from the table using the towel in the robot's gripper. The task is considered finished when the tape is no longer on the table.
    \item Energy Drink Kitchen\\
    $\bullet$ Setup: The robot starts in the user's bedroom near the doorway. An energy drink is placed on the table in the kitchen.\\
    $\bullet$ Description: The user has to drive the robot out of the room, through a hallway, and into the kitchen. The robot has to grab the energy drink and pick up the drink. The task is considered finished when the drink is picked up.
    \item Towel Wipe Face\\
    $\bullet$ Setup: A small black hand towel is placed folded on a table to the left of the user.\\
    $\bullet$ Description: The user has to grab the towel and bring the towel to the user's face with the robot. The task is considered finished when the user wipes their face using the hand towel.
    \item Bring Towel to User\\
    $\bullet$ Setup: A large towel typically used to cushion the user is placed on the bed of the user. \\
    $\bullet$ Description: The user has to bring the towel to the user with the robot. The task is considered finished when the user takes the towel from the robot.  
    \item Aquarium Photo\\
    $\bullet$ Setup: The robot is moved to the right of the aquarium with an iPad Mini placed in the gripper of the robot. The camera app is opened on the iPad mini.  \\
    $\bullet$ Description: The user has to bring the tablet and point it toward the aquarium. The task is considered finished when the user takes a picture of the aquarium using voice commands. 
\end{enumerate}
\subsection{Practice Tasks Setup and Description}
\label{supplementary:sec-practice-tasks}
\begin{enumerate}
    \item Drink from Cup\\
    $\bullet$ Setup: Same as in Section \ref{supplementary:task-setup}.\\
    $\bullet$ Description: Same as in Section \ref{supplementary:task-setup}.
    \item Turn Off Light\\
    $\bullet$ Setup: The robot is placed near the light switch. \\
    $\bullet$ Description: The user has to use the robot to turn off the light. The gripper has to tap the light switch, and the task is considered finished when the light has been turned off. 
    \item Adjust Curtains\\
    $\bullet$ Setup: The curtains are open, allowing light to enter the bedroom. \\
    $\bullet$ Description: The user has to use the robot to grab the curtain and slide the curtain to partially close the curtain. The task is considered finished when the user has grabbed the curtain and moved it any distance to partially close the curtain. 
    \item Open Door\\
    $\bullet$ Setup: The door is closed and the robot is placed near the door. The user was tested with joystick control for only this task. \\
    $\bullet$ Description: The user has to open the door. The task is considered finished when the robot has opened the door and pushed it open to a 45 degree angle. 
\end{enumerate}

\begin{figure}[H]
  \centering
  \includegraphics[width=0.9\linewidth]{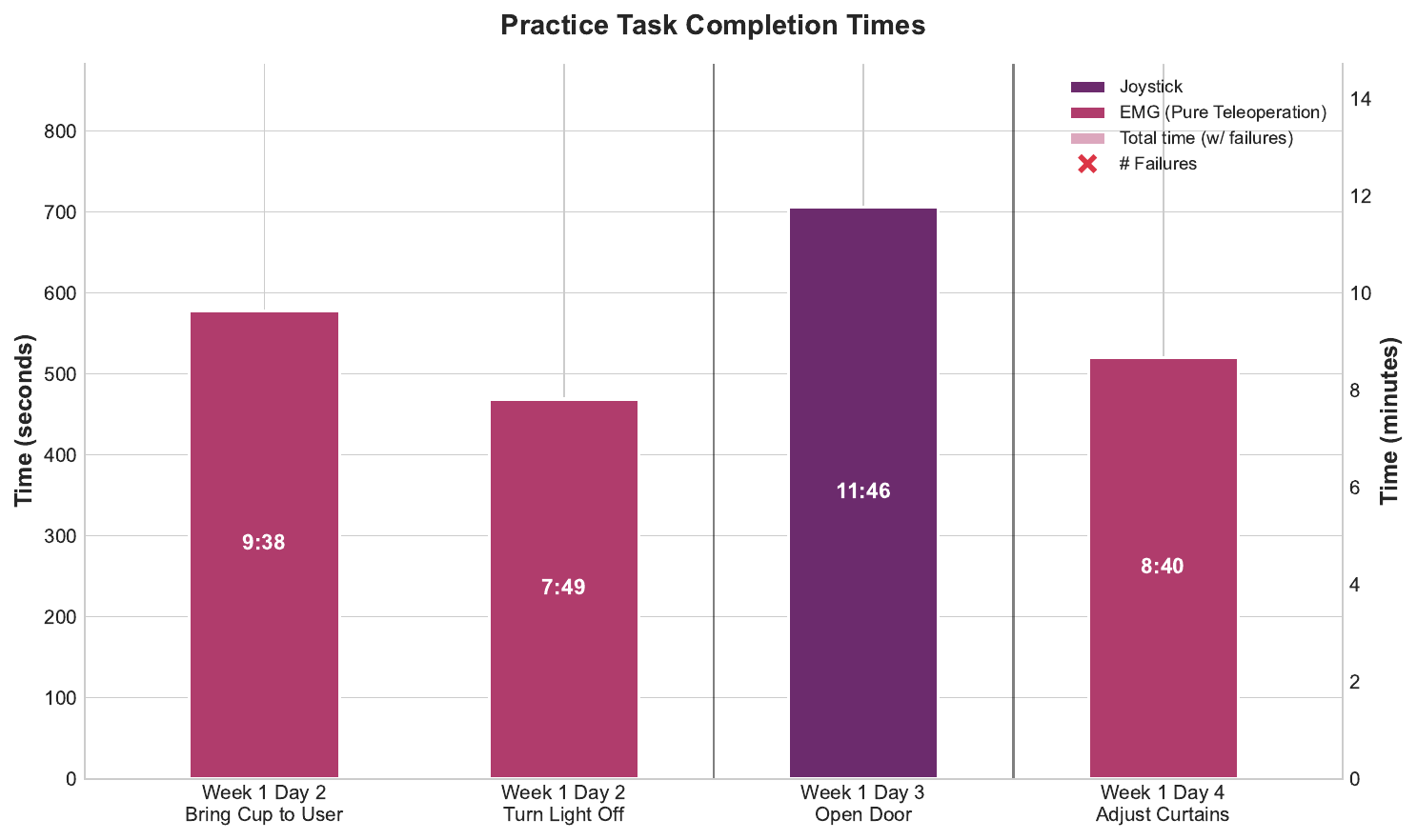}
  \includegraphics[width=0.9\linewidth]{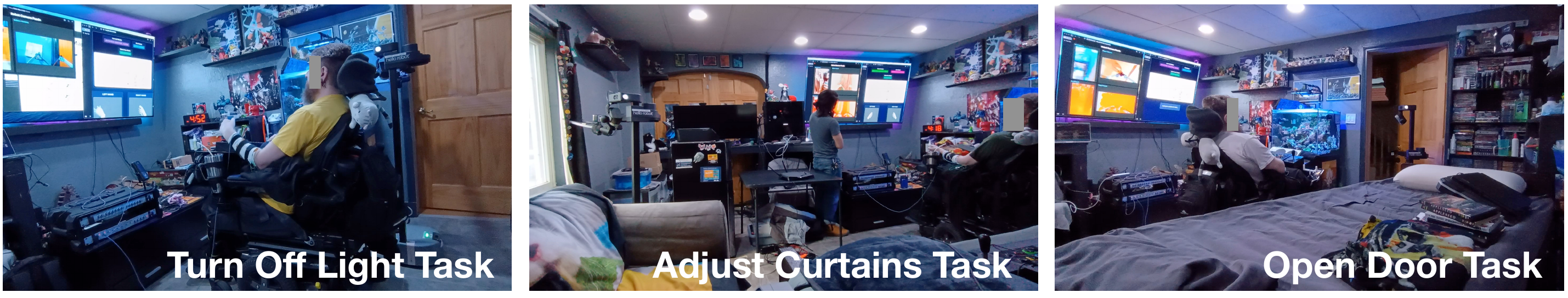}
  \caption{Additional tasks performed in the first week as practice tasks. In this week, we tested using 5-gesture classifiers in each hand, using only RGB camera feeds without a fisheye lens camera, and tested using a joystick for control. We found that the camera feed was also unreliable, but was fixed by switching the internet browser used in the second week.}
  \label{supplementary:practice-tasks}
\end{figure}

We also show additional tasks performed in the first week as practice tasks in Figure~\ref{supplementary:practice-tasks}.

\subsection{Accuracies Across Users for Bimanual EMG Sleeve}
\label{supplementary:accuracies-across-users}

Individual hand accuracies across different preprocessing methods, architectures, and users' hands are presented in Table~\ref{tab:per-hand-cv} for 15 gesture trials. We limit the evaluation to 15 gesture trials for both users because 15 gesture trials is the max number of gesture trials collected for User 2 for both hands. 

\begin{table}[ht]
  \centering
  \scriptsize
  \caption{Per-participant-hand 5-fold CV accuracy (\%) (mean, std across folds). Each row is a model/feature method; columns are the four participant-hands. lin = linear probing, ft = fine-tuning (ResNet18/TinyViT pretrained on ImageNet).}
  \label{tab:per-hand-cv}
  \begin{tabular}{l llll}
    \toprule
    \textbf{Method}
    & \makecell{User 1\\left}
    & \makecell{User 1\\right}
    & \makecell{User 2\\left}
    & \makecell{User 2\\right} \\
    \midrule
    \multicolumn{5}{l}{\textit{RMS}} \\
    MLP                 & 79.7\,(6.1) & 54.1\,(8.8) & 90.5\,(2.7) & 94.8\,(3.4) \\
    CNN                 & \textbf{82.6}\,(7.8) & 55.7\,(7.4) & 93.8\,(1.9) & 94.0\,(5.2) \\
    Transformer         & 72.5\,(4.9) & 53.8\,(6.8) & 70.1\,(5.0) & 76.4\,(4.3) \\
    ResNet18 lin        & 73.8\,(9.3) & 54.1\,(7.0) & 80.8\,(2.7) & 91.8\,(4.5) \\
    ResNet18 ft         & 68.0\,(15.1) & 38.0\,(10.2) & 51.9\,(19.8) & 74.8\,(19.9) \\
    TinyViT lin         & 77.5\,(8.9) & 53.6\,(6.5) & 77.2\,(8.8) & 90.1\,(5.1) \\
    TinyViT ft          & 31.2\,(12.0) & 25.3\,(9.3) & 39.4\,(13.9) & 41.9\,(11.1) \\

    \midrule
    \multicolumn{5}{l}{\textit{Raw}} \\
    MLP                 & 47.4\,(3.5) & 37.4\,(6.3) & 42.5\,(3.7) & 60.0\,(2.9) \\
    CNN                 & 71.4\,(9.3) & 40.9\,(9.9) & 91.9\,(4.8) & 91.5\,(4.2) \\
    TDS                 & 78.7\,(9.6) & 55.0\,(5.8) & 93.0\,(3.9) & 95.6\,(3.3) \\
    Transformer         & 66.4\,(9.2) & 47.0\,(5.8) & 71.8\,(4.8) & 67.4\,(3.1) \\
    MetaCNN-LSTM        & 78.0\,(13.9) & 54.9\,(4.5) & 90.4\,(2.8) & 96.1\,(1.5) \\
    ResNet18 lin        & 72.0\,(5.2) & 53.8\,(6.6) & 83.2\,(7.4) & 91.0\,(2.6) \\
    ResNet18 ft         & 75.4\,(7.1) & 52.5\,(8.2) & 90.3\,(4.7) & 94.6\,(5.4) \\
    TinyViT lin         & 76.2\,(6.0) & 52.7\,(5.2) & 79.4\,(6.5) & 90.9\,(3.0) \\
    TinyViT ft          & 70.4\,(12.2) & 42.3\,(13.7) & 73.1\,(17.5) & 93.1\,(4.4) \\

    \midrule
    \multicolumn{5}{l}{\textit{CWT}} \\
    MLP                 & 81.0\,(4.8) & 53.9\,(4.0) & 94.0\,(2.1) & \textbf{96.6}\,(2.6) \\
    CNN                 & 81.1\,(8.7) & \textbf{57.6}\,(4.7) & \textbf{96.1}\,(1.4) & 95.4\,(2.2) \\
    Transformer         & 60.0\,(5.5) & 44.3\,(1.9) & 80.1\,(1.9) & 93.5\,(2.5) \\
    ResNet18 lin        & 71.2\,(7.8) & 52.2\,(6.5) & 73.1\,(7.1) & 81.9\,(2.8) \\
    ResNet18 ft         & 63.3\,(8.8) & 36.9\,(9.5) & 66.1\,(17.6) & 90.2\,(3.1) \\
    TinyViT lin         & 70.4\,(5.9) & 52.8\,(5.0) & 70.8\,(5.3) & 79.2\,(1.7) \\
    TinyViT ft          & 26.0\,(5.5) & 18.4\,(2.3) & 28.1\,(7.3) & 30.0\,(8.9) \\

    \midrule
    \multicolumn{5}{l}{\textit{MUAP}} \\
    MLP                 & 53.5\,(9.5) & 52.4\,(6.2) & 59.4\,(9.4) & 89.1\,(4.0) \\
    CNN                 & 52.7\,(9.8) & 53.1\,(7.3) & 57.8\,(7.6) & 87.9\,(4.6) \\
    TDS                 & 55.2\,(11.9) & 52.8\,(5.8) & 54.1\,(8.9) & 87.5\,(5.2) \\
    Transformer         & 42.4\,(7.4) & 27.0\,(4.7) & 39.3\,(5.6) & 59.9\,(7.7) \\
    MetaCNN-LSTM        & 53.2\,(9.6) & 52.6\,(3.6) & 54.0\,(11.9) & 87.4\,(4.0) \\
    ResNet18 lin        & 52.8\,(11.3) & 47.9\,(8.8) & 54.0\,(8.1) & 84.7\,(4.0) \\
    ResNet18 ft         & 53.8\,(12.3) & 50.8\,(6.8) & 54.1\,(9.5) & 88.8\,(5.2) \\
    TinyViT lin         & 54.0\,(9.2) & 46.9\,(8.5) & 51.5\,(8.3) & 81.8\,(4.6) \\
    TinyViT ft          & 47.1\,(8.0) & 40.2\,(5.2) & 50.6\,(6.7) & 78.1\,(10.3) \\

    \bottomrule
  \end{tabular}
\end{table}

\subsection{Details on Feature Representations}
\label{supplementary:details-on-preprocessing}

All features are computed from the same 80\,ms (320-sample) analysis window, slid at a 40\,ms stride, over the 128-channel ($8{\times}16$) HD-sEMG electrode grid sampled at 4\,kHz.

\textbf{RMS.} The root-mean-square amplitude of each channel over the window, giving a single value per electrode reshaped to the $8{\times}16$ grid ($1{\times}8{\times}16$). This collapses the time axis to one envelope value per channel and is the standard, lightweight surface-EMG amplitude feature.

\textbf{Raw.} The unprocessed EMG window itself, with channels arranged in electrode order as a $1{\times}128{\times}320$ (channel\,$\times$\,time) array. It discards no information and lets the model learn its own spatial and temporal filters directly from the waveform.

\textbf{CWT.} A continuous wavelet transform applied independently to each channel, producing a time--frequency scalogram (12 frequency scales $\times$ 13 time bins) per channel and stacking all 128 channels into a $128{\times}12{\times}13$ tensor. This exposes the signal's spectral content over time. For the ImageNet-pretrained backbones, the 128 per-channel scalograms are tiled into the physical $8{\times}16$ electrode layout to form a single time--frequency image.

\textbf{MUAP.} A representation of neural-drive obtained by convolutive blind source separation, which decomposes the multichannel EMG into individual motor-unit spike trains, adapted from~\citet{yang2025intuitive}. A separate decomposition bank is fit per gesture on the training trials only (refit within each cross-validation fold to avoid leakage). Each identified unit's spike train is convolved with a Gaussian kernel to yield a smoothed firing-rate (Hz) trace over the window, and all units are stacked into a unit\,$\times$\,time map ($1{\times}U{\times}64$, with $U\approx22$--$30$ units). Unlike the surface-signal features above, this encodes the estimated firing activity of the underlying motor units.

\textbf{MPF.} A spectral-connectivity feature based on inter-channel cross-spectral density (CSD) as described in~\citep{kaifosh2025generic} and adapted to the data format in our work. The grid is first subsampled to 32 channels. For each window, complex CSD matrices are estimated between all channel pairs across frequency, capturing the coherence and phase relationships between the electrodes. These are binned into frequency bands and regularized and symmetrized into symmetric-positive-definite (SPD) matrices. Each SPD matrix is then projected onto the Riemannian tangent space (a Euclidean linearization of the SPD manifold about the training-set mean, fit on training data only), and the off-diagonal entries are extracted as the feature vector.

\subsection{Details on Architectures}
\label{supplementary:details-on-architectures}

\textbf{MLP.} A fully-connected baseline. The input feature is flattened and passed through two hidden layers of 128 and 64 units, each with a GELU nonlinearity and dropout ($p{=}0.5$), followed by a linear classification layer. It treats the feature as an unstructured vector, ignoring spatial or temporal layout.

\textbf{CNN.} A small 2-D convolutional network applied to the feature map. Two convolutional blocks (32 then 64 filters, each a $3{\times}3$ convolution with padding 1, batch normalization, ReLU, and $2{\times}2$ max pooling) extract local spatial features; the result is flattened and passed through a fully-connected head (Linear$\,$128, batch norm, ReLU, dropout $p{=}0.5$, linear to the class logits).

\textbf{Transformer.} A compact vision-transformer-style patch encoder that treats the input feature as a 2-D map. A single $4{\times}4$ convolution with stride 16 acts as a patch embedding, projecting the input into 16-dimensional tokens; four Transformer encoder layers (embedding dim 16, 2 attention heads, feed-forward dim 16, dropout $p{=}0.1$) apply self-attention across the patch tokens; an adaptive average pool and a linear head produce the class logits.

\textbf{TDS.} A fully convolutional sequence encoder adapted from the time-depth-separable architecture of \citet{hannun2019sequence}. The (channel\,$\times$\,time) input passes through an initial 1-D convolutional block (32 channels) followed by two TDS stages, each stacking TDS blocks that factor a spatiotemporal convolution into a time-only depthwise convolution plus a channelwise (pointwise) mixing step. This separation captures temporal structure with far fewer parameters than a dense 2-D convolution; the extracted features are flattened and passed to a small fully-connected head. As a sequence model it requires an explicit time axis and is therefore applied only to the Raw and MUAP features.

\textbf{MetaCNN-LSTM.} A convolutional-recurrent decoder adapting the discrete-gesture model of Meta's neuromotor-interface work~\citep{kaifosh2025generic}. A 1-D temporal convolution (256 channels, kernel 16, stride 8, batch norm, ReLU, dropout $p{=}0.3$) featurizes and downsamples the (channel\,$\times$\,time) input; a 3-layer unidirectional LSTM (hidden size 256) models the temporal dynamics; and the final-timestep hidden state is passed to a linear classifier. Like TDS, it consumes a sequence and is applied only to the Raw and MUAP features.

\textbf{ResNet18.} The standard 18-layer residual network~\citep{he2016deep} ResNet pretrained on ImageNet (torchvision \texttt{IMAGENET1K\_V1} weights), with the final fully-connected layer replaced by a linear layer over the five gesture classes. The single-channel feature is replicated to three channels, bilinearly resized to $224{\times}224$, and ImageNet-normalized (for CWT, the 128 per-channel scalograms are first tiled into the $8{\times}16$ electrode layout to form one image). In linear probing (\texttt{lin}) the backbone is frozen and only the head is trained; in fine-tuning (\texttt{ft}) all weights are updated.

\textbf{TinyViT} A compact 5M-parameter vision transformer~\citep{wu2022tinyvit} (\texttt{tiny\_vit\_5m\_224}, pretrained on ImageNet-22k and fine-tuned on ImageNet-1k via \texttt{timm}), with its classification head replaced by a linear layer over the five classes. It uses the same input adaptation as ResNet18 (replicate to three channels, resize to $224{\times}224$, ImageNet-normalize), and the same linear-probe / fine-tune protocol.

\subsection{Five-Gesture Test Accuracy in First Week}

\begin{figure}[H]
  \centering
  \includegraphics[width=1.0\linewidth]{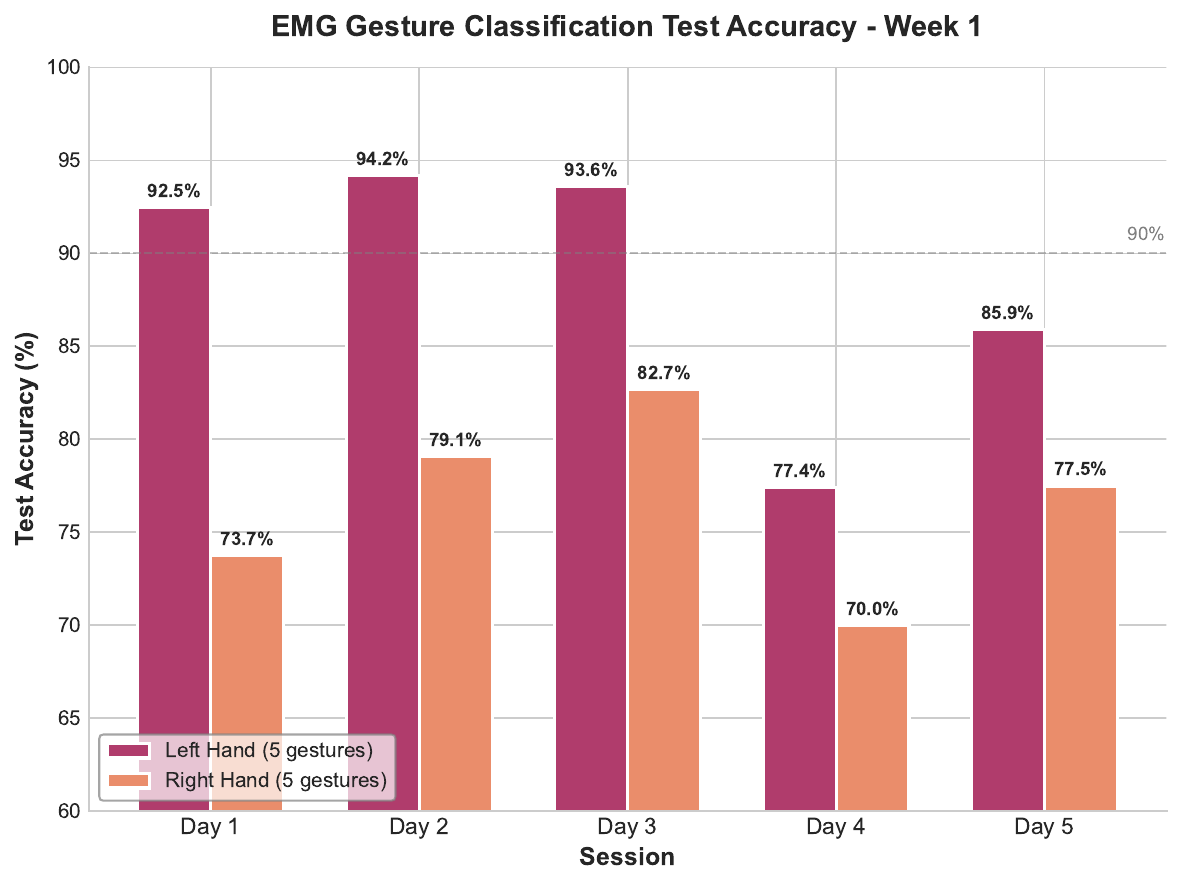}
  \caption{Five-gesture test accuracies for both hands for first week. The right hand shows significantly lower accuracy than the left hand for 5 gestures. }
  \label{supplementary:test-accuracy-week1-5-gestures}
\end{figure}

The test accuracies for the classification of five-gestures were consistently higher on the left hand than on the right during the first five days of the study. Supplementary Fig.~\ref{supplementary:test-accuracy-week1-5-gestures} shows that right-hand test accuracy was lower than left-hand accuracy by a mean of 12.1\% ($\pm 4.3\%$) across these sessions. This asymmetry motivated personalization of the interface by reducing the right-hand vocabulary to three gestures for the finalized real-time control phase in Week~2, after which higher right-hand test accuracies were observed (Fig.~\ref{fig:test-accuracy}).

\begin{figure}[H]
  \centering
  \includegraphics[width=1.0\linewidth]{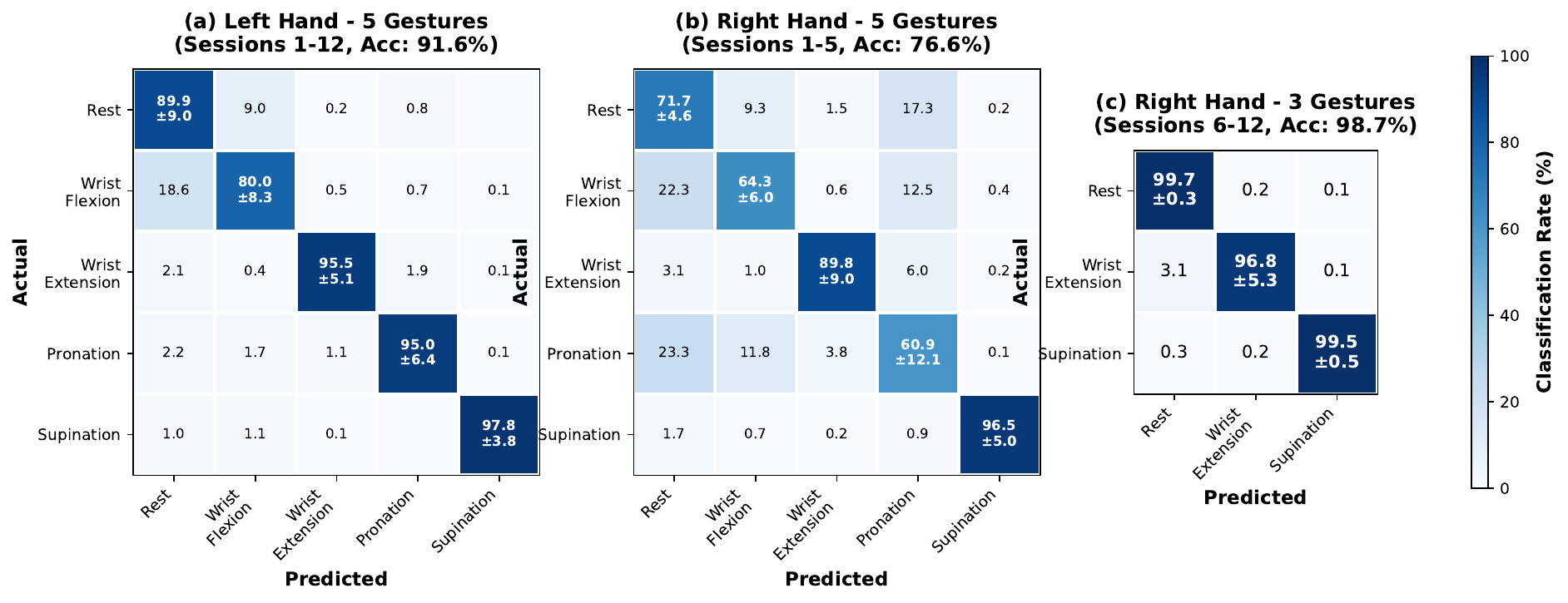}
  \caption{Average confusion matrices for classification for 5 gestures on each hand and 3 gestures on the right hand.}
  \label{supplementary:confusion-matrices}
\end{figure}

The two gestures with the lowest classification performance were consistently wrist flexion and wrist pronation on the right hand (Supplementary Fig.~\ref{supplementary:confusion-matrices}). These gestures also exhibited the smallest observable range of motion among the five gestures, with little to no visible movement during execution.

\subsection{Hardware System Overview}

The bimanual \ac{HDEMG} sleeves worn on the user’s forearms sense neuromotor signals that are used to decode the intended hand gestures of the user. The inner and outer views of the sleeve are shown in Supplementary Figures~\ref{supplementary:sleeve-inner-view}--\ref{supplementary:sleeve-outer-view}. Signals are acquired using an Intan RHD 2164 headstage connected via SPI to an Intan RHD Recording Controller. The recording controller interfaces with a companion laptop, which predicts the intended gesture for each hand in real time. The laptop then sends robot control signals via wireless UDP to a local router, which forwards the final commands to the Stretch robot. Voice commands from the user are registered directly by the microphone on the companion laptop. A high-level hardware overview is shown in the Supplementary Figure~\ref{supplementary:hardware}.

\begin{figure}[H]
  \centering
  \includegraphics[width=0.9\linewidth]{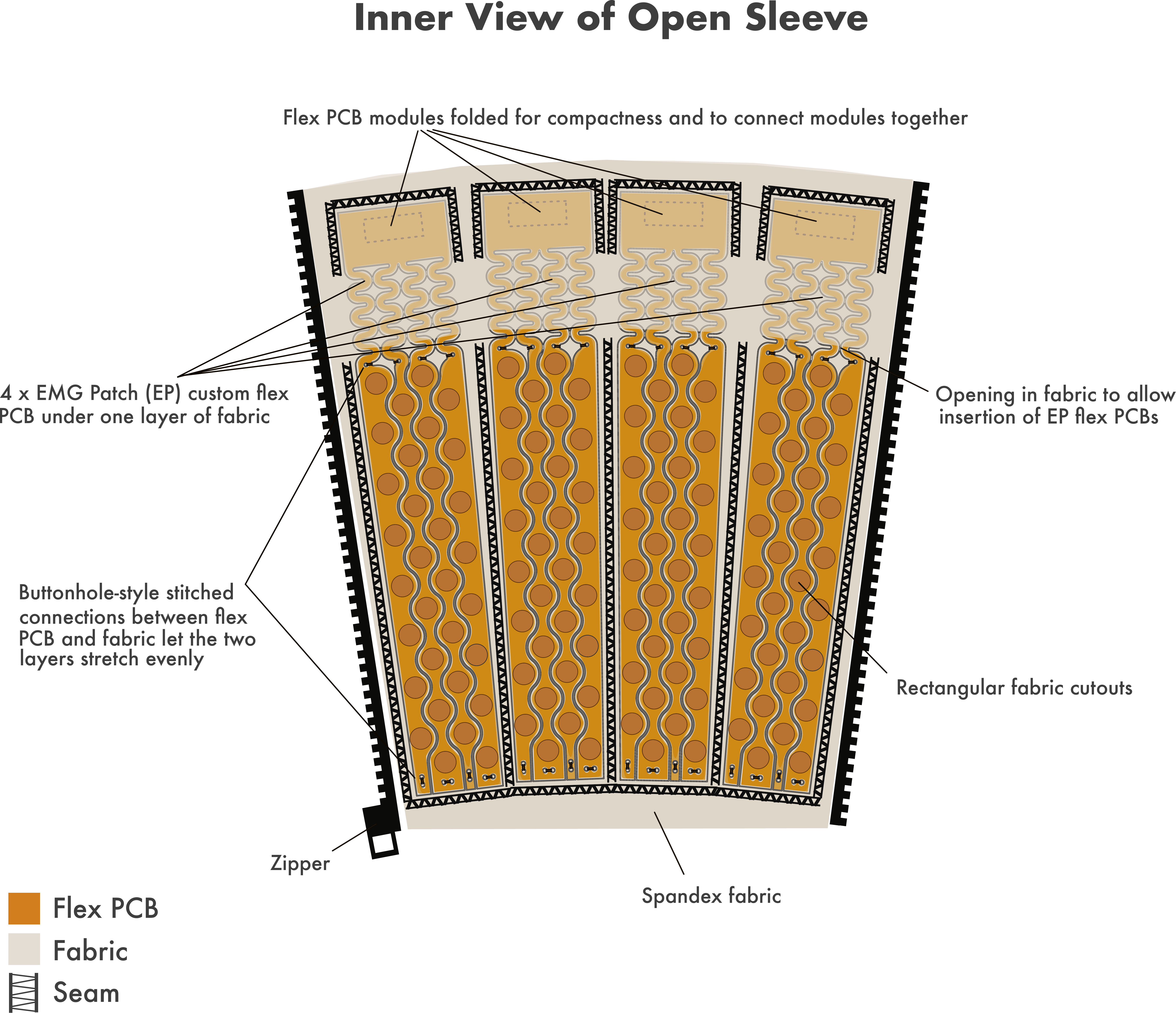}
  \caption{Inner view of diagram for sleeve. The 128 gold-plated electrodes are on this side and make contact with the skin.}
  \label{supplementary:sleeve-inner-view}
\end{figure}
\vspace{-1em}

\begin{figure}[H]
  \centering
  \includegraphics[width=0.75\linewidth]{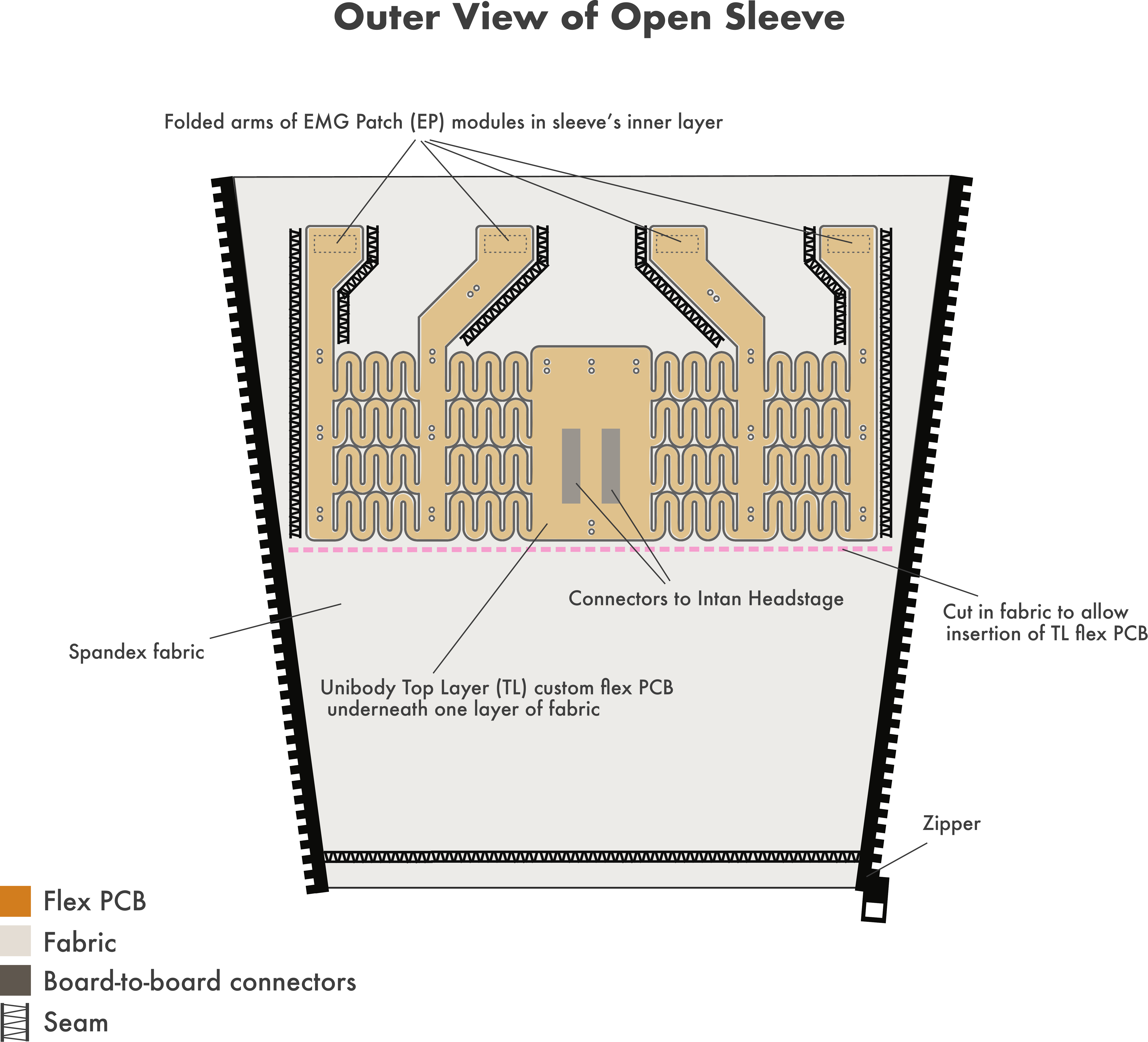}
  \caption{Outer view of diagram for sleeve. The center shows the connectors used to connect to the Intan RHD 2164 headstage. }
  \label{supplementary:sleeve-outer-view}
\end{figure}

\begin{figure}[H]
  \centering
  \includegraphics[width=1.0\linewidth]{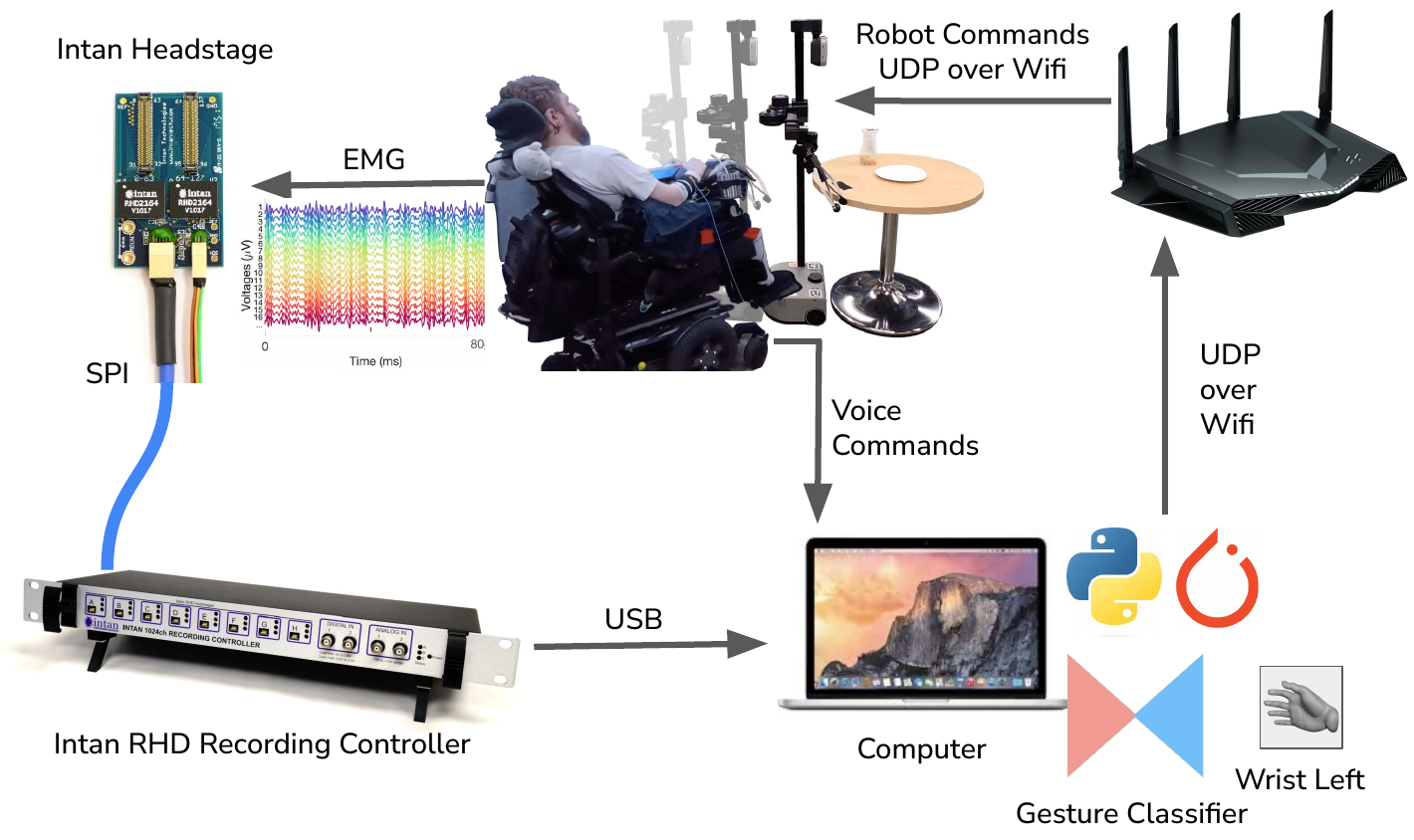}
  \caption{Hardware system overview showing hardware used for EMG and voice control of robot.}
  \label{supplementary:hardware}
\end{figure}

\subsection{Task Times for Teleoperation vs Shared Autonomy}

\begin{figure}[H]
  \centering
  \includegraphics[width=1.0\linewidth]{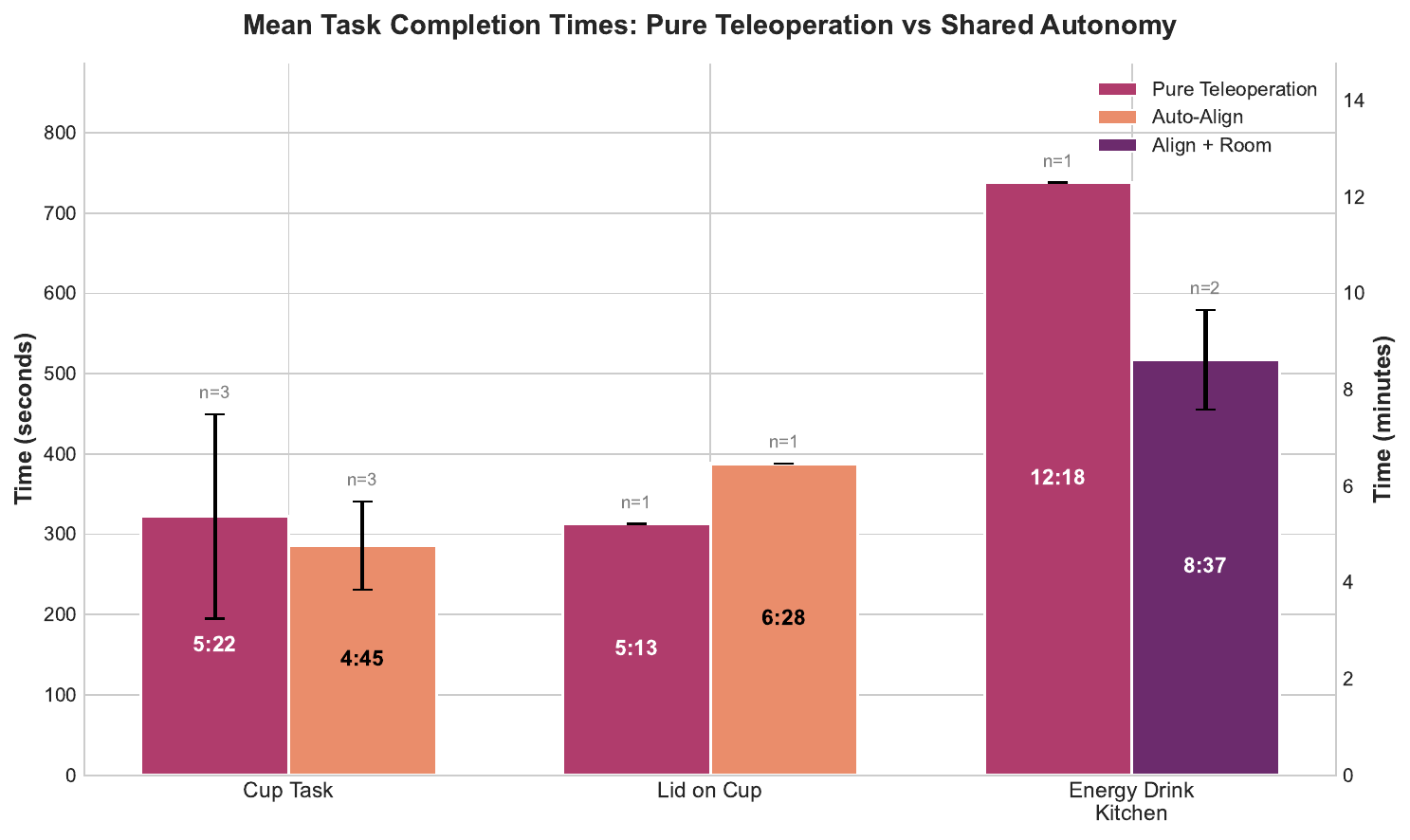}
  \vspace{-7.5mm}
  \caption{Mean and standard deviation task times for all tasks with shared autonomy. We find higher task times for tasks in which auto-align is practiced only once, but lower mean task times for tasks in which the user used shared autonomy multiple times.}
  \label{supplementary:task-time-comparisons-all-shared-autonomy}
  \vspace{-4.5mm}
\end{figure}

Task times for teleoperation and shared autonomy have a lower mean for shared autonomy for two out of the three tasks (Fig.~\ref{supplementary:task-time-comparisons-all-shared-autonomy}. We find that although shared autonomy may benefit users in simplifying alignment during manipulation and navigation, the learning required for the user to use shared autonomy properly for a specific task may increase task times during the first trial of a task. More evaluations are needed to conclude the benefits of shared autonomy in a home deployment setting, especially based on the user's ability to adapt to new tasks after gaining familiarity. 
\vspace{-2mm}

\subsection{Default Joint Speeds}
We list the default robot joint speeds in Table~\ref{tab:robot_joint_speeds}. The base turning speeds up by \(4\times\) after 3 seconds rather than by \(2\times\) like the other joints to allow precise rotations during the slow condition and efficient rotation during the fast condition. 
\begin{table}[ht]
\centering
\vspace{-1.5mm}
\caption{Default robot joint speeds in the slow (first 3\,s) and fast (after 3\,s)
phases. Wrist, gripper, and base rotation are angular (rad/s); base trans.
and arm joints are linear (m/s).}
\vspace{-3.5mm}
\label{tab:robot_joint_speeds}
\begin{tabular}{llcc}
\toprule
Joint group & Unit & Slow ($<$3\,s) & Fast ($>$3\,s) \\
\midrule
Wrist (yaw/pitch/roll) & rad/s & 0.6250 & 1.2500 \\
Gripper                & rad/s & 0.3125 & 0.6250 \\
Base -- turn           & rad/s & 0.3125 & 1.2500 \\
\midrule
Base -- drive          & m/s   & 0.313  & 0.625  \\
Arm -- lift            & m/s   & 0.156  & 0.313  \\
Arm -- extend (telescope) & m/s & 0.039 & 0.078  \\
\bottomrule
\end{tabular}
\vspace{-7.5mm}
\end{table}

\subsection{Voice Commands State Machine}
\begin{figure}[H]
  \centering
  \includegraphics[width=0.6\linewidth]{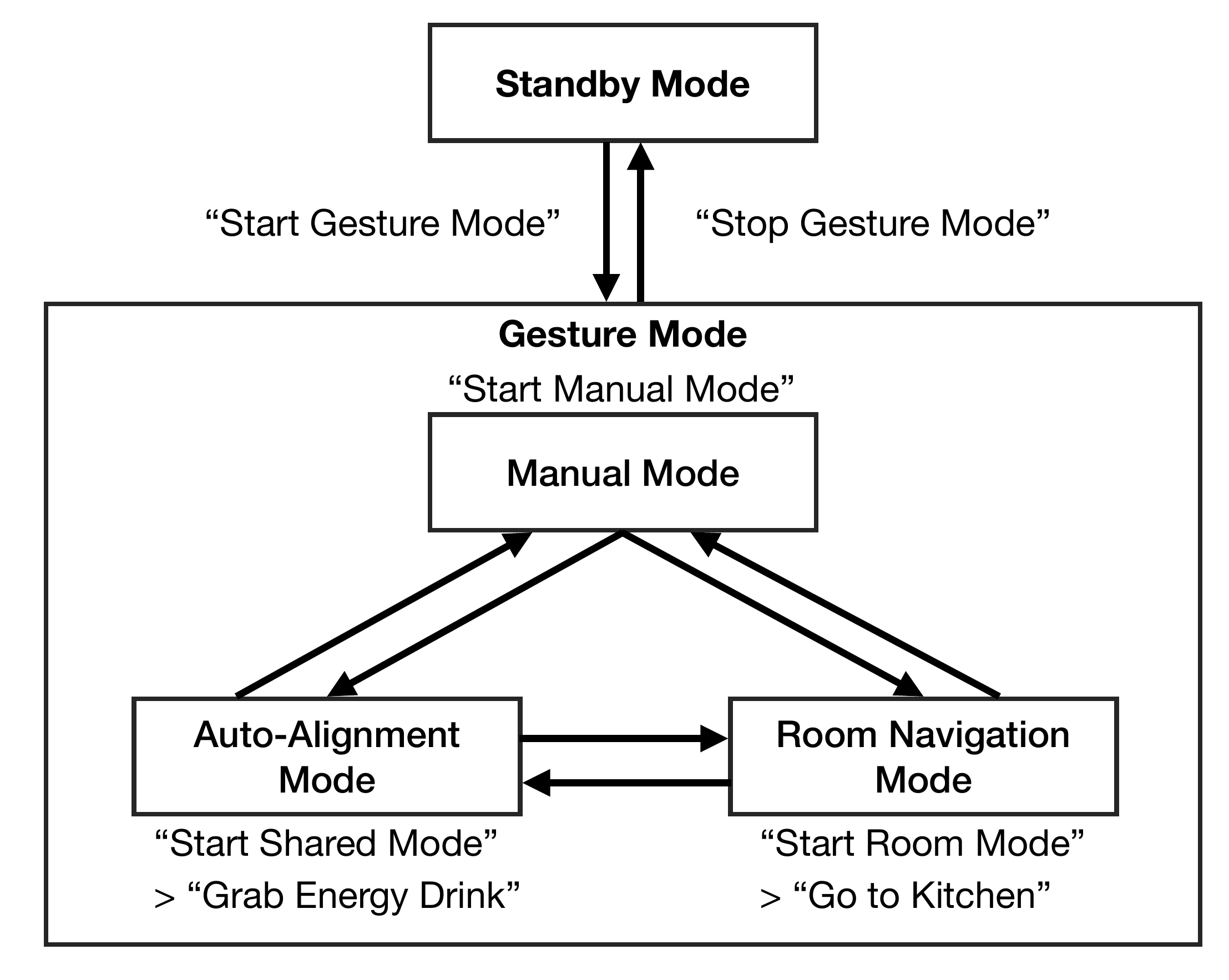}
  \vspace{-3.5mm}
  \caption{Diagram showing different modes that are switched between using voice commands.}
  \vspace{-3.5mm}
  \label{supplementary:voice-switching-modes}
\end{figure}

The modes used for voice commands are in Figure~\ref{supplementary:voice-switching-modes}.

\subsection{Experiment on Pre-Training from Other Days and Fine-Tuning on Current Day}
Shown in Table~\ref{tab:xday_lodo_12day}, we find that pretraining + fine-tuning performs only marginally better than training only from data from the same day for the left hand and performs worse for the right hand. We report classifier test performance from training only with data from the same day in the \textit{same-day} column, from training only on all other days in the \textit{zero-shot} column, and fine-tuning with data from the same day after pretraining with data from all other days in the \textit{fine-tuning} column.

\begin{table}[ht]
  \centering
  \caption{Cross-day performance with 5 within-day cross-validation accuracy (\%) with the full 12-days (User 1, reduced to 30
  trials/gesture/day, mean\,$\pm$\,std over 5 within-day folds).}
  \label{tab:xday_lodo_12day}
  \vspace{-3.5mm}
  \begin{tabular}{llccc}  
  \toprule  
  Hand & Day & \textit{same-day} & \textit{zero-shot} & \textit{+fine-tuning} \\  
  \midrule
  \multirow{13}{*}{Left (5 gest.)}
   & 1  & $85.2\pm5.2$ & $74.0\pm6.9$ & $86.9\pm5.2$ \\
   & 2  & $92.3\pm2.5$ & $86.1\pm4.5$ & $93.8\pm2.1$ \\
   & 3  & $94.8\pm3.3$ & $94.4\pm1.6$ & $97.5\pm1.0$ \\
   & 4  & $77.3\pm3.5$ & $73.4\pm4.9$ & $78.8\pm3.7$ \\
   & 5  & $83.1\pm4.8$ & $78.1\pm4.3$ & $84.3\pm4.4$ \\
   & 6  & $98.3\pm0.6$ & $95.2\pm2.2$ & $99.0\pm0.4$ \\
   & 7  & $95.9\pm2.3$ & $88.8\pm3.9$ & $96.8\pm2.1$ \\
   & 8  & $97.5\pm1.2$ & $93.5\pm3.9$ & $98.4\pm0.7$ \\
   & 9  & $94.5\pm2.3$ & $93.5\pm2.3$ & $96.1\pm1.4$ \\
   & 10  & $88.8\pm6.0$ & $89.5\pm4.5$ & $89.8\pm5.7$ \\
   & 11 & $84.9\pm7.5$ & $87.5\pm8.9$ & $87.8\pm8.0$ \\
   & 12 & $77.0\pm1.4$ & $70.9\pm6.8$ & $81.9\pm2.7$ \\
   & \textbf{mean} & $\mathbf{89.1}$ & $\mathbf{85.4}$ & $\mathbf{90.9}$ \\
  \midrule
  \multirow{13}{*}{Right (3 gest.)}
   & 1  & $91.9\pm6.6$  & $74.8\pm9.2$  & $86.3\pm10.0$ \\
   & 2  & $97.8\pm2.2$  & $95.2\pm2.1$  & $98.6\pm0.7$ \\
   & 3  & $99.6\pm0.3$  & $75.5\pm4.5$  & $99.6\pm0.2$ \\
   & 4  & $90.4\pm5.1$  & $85.9\pm7.1$  & $90.1\pm5.4$ \\
   & 5  & $95.7\pm3.7$  & $96.1\pm3.8$  & $96.3\pm3.6$ \\
   & 6  & $99.0\pm0.8$  & $56.3\pm9.5$  & $99.2\pm1.1$ \\
   & 7  & $97.4\pm1.5$  & $89.3\pm4.9$  & $96.9\pm1.9$ \\
   & 8  & $99.2\pm1.3$  & $96.8\pm1.8$  & $99.8\pm0.2$ \\
   & 9  & $99.8\pm0.1$  & $98.7\pm0.5$  & $100.0\pm0.1$ \\
   & 10  & $93.9\pm6.2$  & $92.4\pm6.8$  & $94.9\pm5.0$ \\
   & 11 & $99.4\pm0.8$  & $97.7\pm1.6$  & $99.8\pm0.3$ \\
   & 12 & $94.7\pm1.9$  & $90.6\pm5.0$  & $94.8\pm2.3$ \\
   & \textbf{mean} & $\mathbf{96.6}$ & $\mathbf{87.4}$ & $\mathbf{96.4}$ \\
  \bottomrule
  \end{tabular}
  \vspace{-7.5mm}
  \end{table}

\subsection{Human Participants Research}
Before inclusion, participants gave their written informed consent, including photography and video, and agreed that this material can be used in journals and other public media. Participants were above the age of 18 and recruited through the use of fliers and email. The study protocol was approved by the Carnegie Mellon University Institutional Review Board, protocol 2026.00000012.

\subsection{Code, Hardware Design, and Data Availability}
We plan to release all code for EMG classification, real-time robot control, designs and fabrication guide for hardware, and EMG data from users upon acceptance of the manuscript. 



\end{document}